\documentclass{article}
\usepackage{arxiv}

\usepackage[utf8]{inputenc} 
\usepackage[T1]{fontenc}    
\usepackage{hyperref}       
\usepackage{url}            
\usepackage{booktabs}       
\usepackage{amsfonts}       
\usepackage{nicefrac}       
\usepackage{microtype}      
\usepackage{lipsum}		
\usepackage{graphicx}
\usepackage{natbib}
\usepackage{doi}

\usepackage{tabularx}
\usepackage{booktabs}
\usepackage{array}
\usepackage{amsmath}
\usepackage[table]{xcolor}
\usepackage{placeins}
\usepackage{caption}
\title{Implementation and Software-in-the-Loop Evaluation of a Trans-Domain Digital Twin Architecture for Autonomous Guidance of the Phoenix VTOL UAV Without Online GNSS}

\author{
	\href{https://orcid.org/0009-0002-1894-8089}{\includegraphics[scale=0.06]{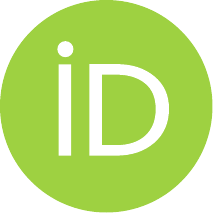}\hspace{1mm}Mansoorali Amiri} \\
	Quebec, Canada \\
}

\renewcommand{\shorttitle}{}

\hypersetup{
pdftitle={Implementation and Software-in-the-Loop Evaluation of a Trans-Domain Digital Twin Architecture for Autonomous Guidance of the Phoenix VTOL UAV Without Online GNSS},
pdfsubject={GNSS-Denied Autonomous UAV Guidance Using a Trans-Domain Digital Twin Architecture},
pdfauthor={Mansoorali Amiri},
pdfkeywords={Trans-Domain Digital Twin, GNSS-Denied UAV Navigation, Autonomous UAV Guidance, Software-in-the-Loop Simulation, Multi-Sensor State Estimation, Closed-Loop Control},
}

\begin{document}
\maketitle

\begin{abstract}
Autonomous UAV guidance in emergency response, infrastructure monitoring, and cargo transportation requires reliable navigation. However, satellite-based positioning may become unavailable or unreliable in enclosed spaces, blind spots caused by natural obstacles and structures, and environments affected by radio interference and signal spoofing during electronic warfare. This study designs, implements, and evaluates a system based on a Trans-Domain Digital Twin (TDDT) for autonomous guidance of the Phoenix VTOL UAV. Independent navigation, flight dynamics, and meteorological models are coordinated within a cycle of state estimation, prediction, decision-making, command execution, and feedback. Data from inertial, magnetic, pressure, and airspeed sensors, together with environmental information, form the basis for estimation and prediction. Furthermore, a shared state, temporal synchronization, preservation of information provenance, and command traceability establish continuity between models and decisions. The system was tested with a simulated physical twin in a software-in-the-loop environment. Vertical takeoff, movement toward the destination, landing, and autonomous stopping were recorded, and the dynamic model was adjusted to a limited extent based on sensor-derived feedback. Separating the evaluation reference from navigation inputs enabled independent assessment of positioning error and computational performance. The results confirm operational coordination, decision traceability, and closed-loop guidance under the tested conditions, providing a reproducible platform for the development and evaluation of navigation without satellite-based positioning.
\end{abstract}


\section{Introduction}

Autonomous unmanned aerial vehicles (UAVs) require reliable navigation for emergency response, infrastructure monitoring, and cargo transport. However, obstruction, interference, and spoofing of satellite signals make GNSS\footnote{Global Navigation Satellite System}-dependent guidance unreliable. Under these conditions, estimation drift, absolute-position ambiguity, and computational constraints are major challenges \cite{ref01_jarraya2025gnss}\cite{ref02_alqudsi2027state}. Guidance evaluation must also encompass motion reconstruction, position validity, and the consequences of executed commands.

Although multisensor fusion can improve state estimation and resilience to sensor degradation \cite{ref03_ye2023review}, the consistency of a relative trajectory does not establish the accuracy of absolute positioning or confirm arrival at the destination \cite{ref05_maharmeh2025survey}. Calibration errors, data asynchrony, and computational costs also limit estimation accuracy \cite{ref04_lee2025mins}, while the acceptance of invalid data or abrupt switching between sources may disrupt the estimated state and control continuity \cite{ref06_aljaafreh2026integrity}\cite{ref07_zhang2026detection}.

Accordingly, the problem extends beyond sensor fusion: navigation, flight dynamics, the environment, and control must be coordinated within a shared operational cycle. Co-simulation of heterogeneous digital twins enables such interaction \cite{ref08_gil2026model}, while the seven-layer Trans-Domain Digital Twin (TDDT) architecture integrates shared state, cross-domain exchange, decision-making, and feedback \cite{ref09_amiri2026transdomain}. Nevertheless, this coordination does not, in itself, resolve observability limitations or absolute-position errors.

This study implements TDDT for the guidance of the dual-rotor Phoenix VTOL \footnote{Vertical Take-Off and Landing} UAV \footnote{Unmanned Aerial Vehicle} \cite{ref10_wu2019phoenix} without online GNSS. To this end, navigation, flight dynamics, and meteorological twins are connected to ArduPilot/JSBSim within a cycle of estimation, prediction, decision-making, command execution, and feedback. Software-in-the-loop evaluation uses an independent reference to assess autonomous execution of mission phases, navigation accuracy, cross-domain exchange, and processing time, while distinguishing experimental evidence, theoretical capabilities, and simulation limitations.

\subsection{Problem Statement}

In GNSS-denied navigation, estimation drift and the absence of an independent reference limit the validity of absolute positioning \cite{ref01_jarraya2025gnss}\cite{ref02_alqudsi2027state}. Calibration errors, insufficient observability, and asynchronous data sources can produce inconsistent estimates \cite{ref01_jarraya2025gnss}\cite{ref04_lee2025mins}\cite{ref11_lisus2023know}. Even an accurate relative trajectory or a small mean error does not guarantee navigation integrity or actual arrival at the destination \cite{ref11_lisus2023know}\cite{ref05_maharmeh2025survey}.

Errors interact with environmental changes, flight dynamics, and computational delays \cite{ref02_alqudsi2027state}. However, evaluating sensors, models, or controllers separately does not reveal their mutual consequences \cite{ref02_alqudsi2027state}\cite{ref04_lee2025mins}. Therefore, the central problem is insufficient coordination among components and the unreliable propagation of data validity and uncertainty into closed-loop decision-making \cite{ref02_alqudsi2027state}\cite{ref04_lee2025mins}\cite{ref05_maharmeh2025survey}.

\subsection{Research Gap}

First, although co-simulation of heterogeneous digital twins \cite{ref08_gil2026model} and multisensor state estimation \cite{ref01_jarraya2025gnss} have advanced, sufficient evidence is lacking regarding consistent state transfer among navigation, flight dynamics, and the environment and its impact on flight decisions under GNSS-denied conditions \cite{ref08_gil2026model}\cite{ref01_jarraya2025gnss}.

Second, accurate relative motion does not imply valid absolute positioning. Limited observability, initialization errors, and data asynchrony \cite{ref03_ye2023review}\cite{ref04_lee2025mins}, together with unknown data correlations \cite{ref13_forsling2024quarter}, reduce fusion reliability. Therefore, the propagation of data provenance, validity, and uncertainty to the decision-maker still requires evaluation.

Third, despite existing methods for asynchronous data fusion \cite{ref04_lee2025mins}, suspicious-data management, and source recovery \cite{ref07_zhang2026detection}\cite{ref06_aljaafreh2026integrity}, sufficient evidence is lacking regarding the management of data freshness, latency, error-propagation containment, and safe command correction within a multidomain loop \cite{ref04_lee2025mins}\cite{ref07_zhang2026detection}\cite{ref06_aljaafreh2026integrity}.

Fourth, separate evaluation of an estimator or detector does not fully capture the consequences of executed commands and flight feedback \cite{ref07_zhang2026detection}. Likewise, mean accuracy does not establish navigation integrity or confirm arrival at the destination \cite{ref05_maharmeh2025survey}. Therefore, end-to-end validation of system integration and its operational reliability is essential \cite{ref08_gil2026model}\cite{ref07_zhang2026detection}\cite{ref06_aljaafreh2026integrity}\cite{ref05_maharmeh2025survey}.

\subsection{Research Objectives}

The objective of this research is to design and implement TDDT-based autonomous guidance for the Phoenix UAV without online GNSS or human pilot intervention. This objective encompasses multisensor state estimation, distinguishing relative from absolute positioning, drift management, and coordination of navigation, flight dynamics, and meteorological twins through a time-stamped shared state with explicit data provenance.

Furthermore, the research aims to manage source validity and uncertainty, monitor inconsistencies and error propagation, examine dynamic network reconfiguration, select constrained commands, autonomously manage mission phases, and close the estimation, execution, and feedback loop. Navigation accuracy, destination error, control continuity, latency, mission success, and observability limitations are assessed using independent evidence without assuming guaranteed absolute positioning. Proposed capabilities remain distinguished from verified results.

\subsection{Main Contributions and Innovations}

The architectural contribution is the limited operationalization of the seven-layer TDDT structure for Phoenix guidance without online GNSS. The MESA-NAV\footnote{Multi-field Excited Self-Anchoring Navigation}, WingMPET\footnote{Wing Mechanics–Path–Energy–Time Formalism}, and MicroMet twins were integrated through an atomic shared state, time synchronization, preservation of data provenance, and traceable identifiers, establishing operational links among estimation, prediction, and decision-making within coordinated cycles.

The implementation contribution is the direct connection of decisions to ArduPilot/JSBSim and the closure of the command, observation, and limited dynamics-model correction loop. Autonomous mission phases were executed during the tests. Separating simulator ground truth from navigation inputs enabled independent evaluation of destination error, processing time, and exchange validity, while documentation and execution evidence support reproducibility. Nevertheless, the fully dynamic Dynamic N-Twin network, guaranteed absolute positioning, and comprehensive safety were not experimentally validated in this study.

\section{Theoretical Foundations and Related Research}

This section reviews the characteristics of Phoenix, GNSS-independent navigation, and the position of TDDT within related research.

\subsection{Structure and Characteristics of the Phoenix VTOL UAV}

\begin{figure}[htbp]
    \centering
    \begin{minipage}[c]{0.48\textwidth}
        \centering
        \includegraphics[width=\linewidth]{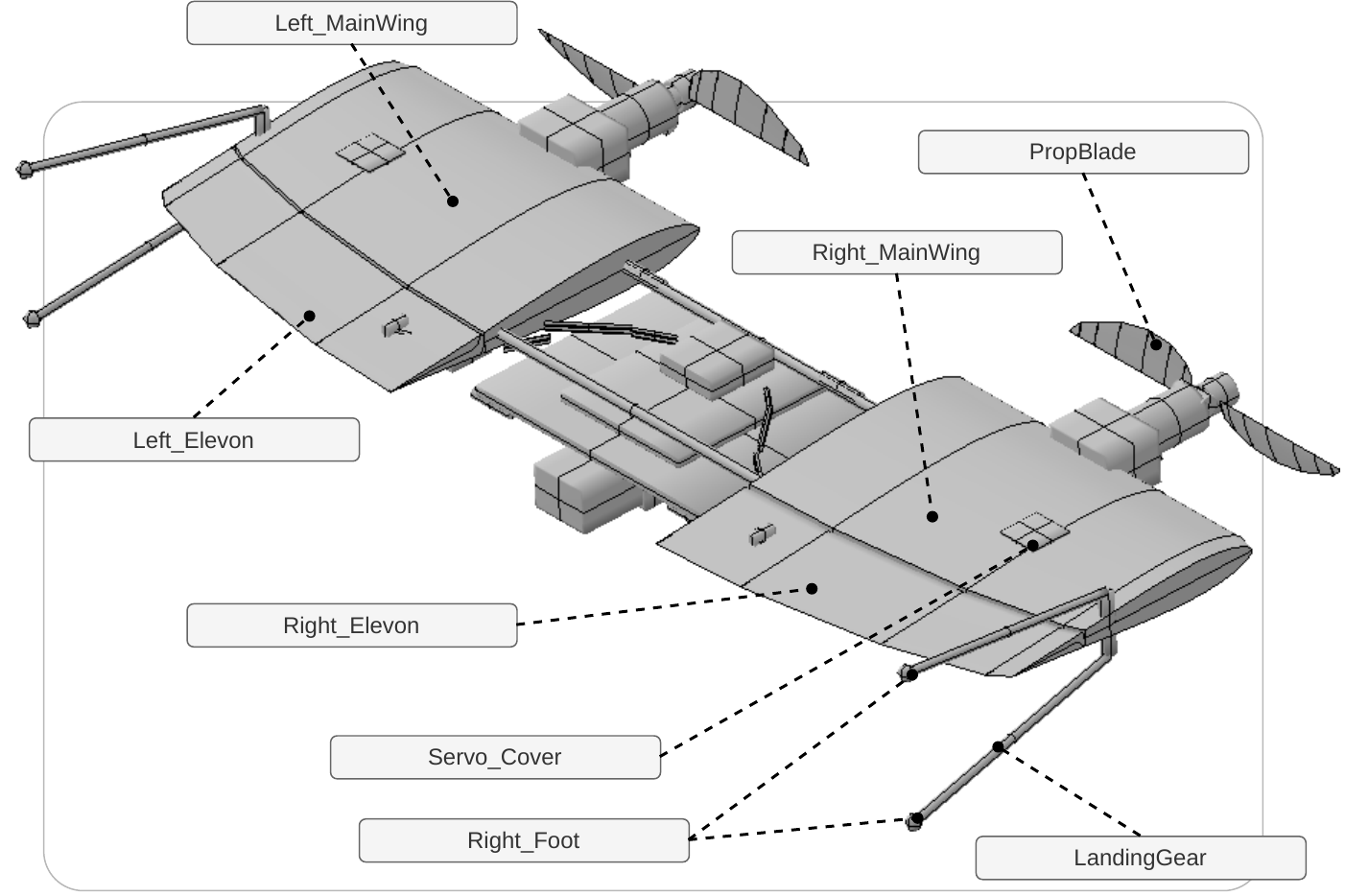}
    \end{minipage}
    \hfill
    \begin{minipage}[c]{0.48\textwidth}
        \centering
        \includegraphics[width=\linewidth,alt={Two annotated technical views of the Phoenix dual-rotor tail-sitter UAV, identifying its wings, elevons, motors, propellers, flight controller, central frame, battery, and landing components.}]{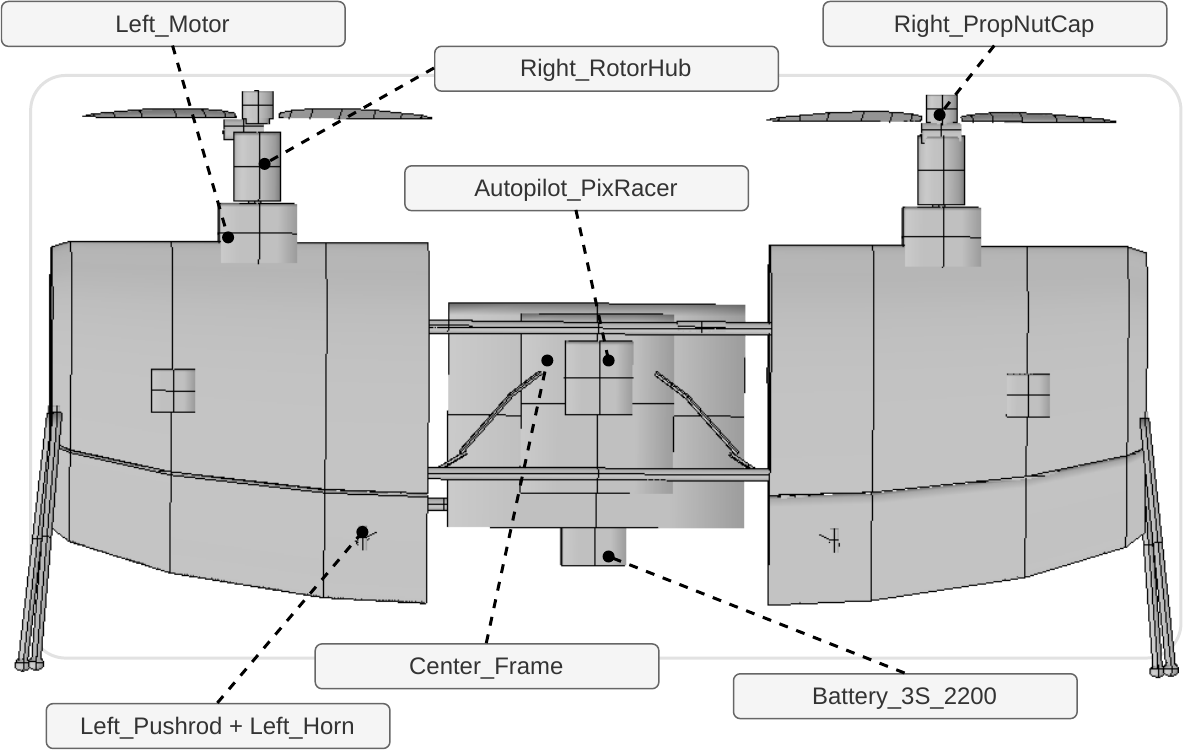}
    \end{minipage}
    \caption{Structural configuration of the Phoenix VTOL dual-rotor tail-sitter. Annotated views identify the main wings, counter-rotating propellers, elevons, flight controller, battery, central frame, and landing components. These elements define the principal propulsion, aerodynamic, and control features of the modeled aircraft. \cite{ref10_wu2019phoenix}}
    \label{fig:drone_comparison}
\end{figure}

Phoenix (Phoenix VTOL) is an open-source, dual-motor tail-sitter UAV capable of vertical takeoff and fixed-wing flight \cite{ref10_wu2019phoenix}, equipped with two counter-rotating propellers and two elevons. During hover, thrust supports the vehicle's weight, whereas in forward flight, the wing generates lift. Transition occurs through a change in body attitude without rotating the motors; the elevons remain effective in the propeller slipstream even at near-zero airspeed \cite{ref10_wu2019phoenix}. The structural configuration and principal propulsion and control components of the Phoenix VTOL are illustrated in Figure \ref{fig:drone_comparison}.

The six-degree-of-freedom model describes motion resulting from gravity, thrust, and aerodynamic forces and moments. Thrust and reaction torque depend on propeller rotational speed, whereas lift, drag, and pitching moment depend on the propeller slipstream and elevon deflection. The wing uses a symmetric Eppler E168 airfoil; the coefficients were identified mainly in hover, and the forward-flight model has lower validity \cite{ref10_wu2019phoenix}.

The reference cascaded controller comprises position, attitude, and angular-rate control. An inverse model converts desired forces and moments into motor and elevon commands, while motor-speed feedback improves the response. Tests have been largely limited to hover; therefore, transition and fixed-wing flight require independent validation \cite{ref10_wu2019phoenix}.

\subsection{Autonomous Navigation Without Reliance on Satellite Positioning}

Navigation without GNSS requires estimation of position, velocity, and orientation. Inertial integration estimates relative displacement, but noise and bias cause drift \cite{ref03_ye2023review}\cite{ref01_jarraya2025gnss}. Learning-based methods can improve velocity estimation but do not determine global coordinates\footnote{Global coordinates denote a fixed Earth-referenced position comprising longitude, latitude, and altitude, independent of the origin of the UAV's motion.} \cite{ref15_bajwa2024dive}. Absolute correction depends on a georeferenced reference; terrain or magnetic matching requires a valid map, spatial distinctiveness, and appropriate initialization \cite{ref01_jarraya2025gnss}\cite{ref16_yao2024gnss}\cite{ref17_zhu2026geomagnetic}. Thus, consistency of the relative trajectory does not guarantee the accuracy of the global trajectory\footnote{The global trajectory is the sequence of UAV positions relative to the geographical origin and destination in a fixed Earth-referenced coordinate system, rather than merely relative displacement.}.

Multisensor fusion is performed using Kalman filters or nonlinear estimators; the magnetometer constrains heading, the barometer constrains altitude, and the Pitot tube constrains airspeed magnitude \cite{ref03_ye2023review}\cite{ref18_deoliveira2024pitot}. Pitot and inertial measurements improve velocity and orientation estimation under observable conditions \cite{ref18_deoliveira2024pitot}. Three-dimensional wind is useful for estimating ground speed, but propeller slipstream, sensor installation, and environmental errors limit its accuracy \cite{ref19_soltaninezhad2025wind}\cite{ref20_chen2026wind}; asynchrony and incorrect calibration cause inconsistency \cite{ref04_lee2025mins}.

Observability is the ability to identify states from measurements. Without an independent reference, global translation\footnote{Global translation is the displacement of the entire estimated trajectory by a constant vector in the Earth-referenced coordinate system, without changing the trajectory's shape or relative displacements.} and some biases remain ambiguous; motion excitation also affects estimation \cite{ref18_deoliveira2024pitot}\cite{ref11_lisus2023know}. Improper linearization can create false confidence \cite{ref11_lisus2023know}. Sharing data between models also creates hidden correlations and double-counting of information; conservative fusion mitigates this risk at the cost of accuracy or availability \cite{ref13_forsling2024quarter}.

Drift, wind, delay, and sensor failure affect estimation quality \cite{ref01_jarraya2025gnss}\cite{ref19_soltaninezhad2025wind}\cite{ref20_chen2026wind}. Graph-based optimization makes use of historical information, but entails computational costs and limitations related to the window and initialization \cite{ref21_lu2026graph}. Moreover, accepting suspect data or abruptly changing sources can disrupt estimation and control \cite{ref07_zhang2026detection}\cite{ref06_aljaafreh2026integrity}.

Without cameras, lidar, radar, or external infrastructure, environmental localization depends on distinctive magnetic or terrain maps \cite{ref16_yao2024gnss}\cite{ref17_zhu2026geomagnetic}. Nevertheless, a low mean error does not establish actual arrival at the destination; evaluation requires error bounds, alert probability, and assessment of navigation integrity \cite{ref05_maharmeh2025survey}. Position validity must be assessed using independent evidence within the limitations of the sensors and mission \cite{ref01_jarraya2025gnss}\cite{ref05_maharmeh2025survey}.

\subsection{Trans-Domain Digital Twin Architecture}

TDDT coordinates heterogeneous twins at the levels of state, error, objective, constraint, and decision \cite{ref09_amiri2026transdomain}. UAV guidance is a cyber-physical system linking software, sensors, actuators, and motion \cite{ref30_sun2026eadr}; navigation, dynamics, and environmental twins contribute to joint prediction through their independent models and update rates \cite{ref09_amiri2026transdomain}\cite{ref22_amiri2026greenhouse}.

The architecture has seven layers: (1) the physical system and its constraints; (2) specialized models; (3) shared state, coupling, and temporal synchronization; (4) prediction and evaluation of alternatives; (5) selection of a constrained control action; (6) execution of the decision in the operational environment; and (7) return of observed errors for model correction \cite{ref09_amiri2026transdomain}. The first and sixth layers are structural and execution views, respectively.

The Shared Trans-Domain State\footnote{The Shared Trans-Domain State (STD) is an integrated and synchronized representation of information from specialized twins in terms of time, state, units, quality, and uncertainty, forming the basis for inter-domain coupling, prediction, and joint decision-making.} aligns the time, units, semantics, quality, provenance, and uncertainty of twin outputs \cite{ref09_amiri2026transdomain}. Six types of coupling---data, model, state, error, objective, and control---propagate cross-domain effects. Dynamic N-Twin proposes dynamic topology, trust weights, coupling strength, and a consistency center, but has not been fully implemented; internal consistency also does not guarantee physical correctness \cite{ref09_amiri2026transdomain}.

Multirate coordination has three loops: a fast loop for response, an intermediate loop for aggregation and guidance, and a slow loop for revising objectives and policies \cite{ref09_amiri2026transdomain}. The TDOC \footnote{Trans-Domain Orchestration Core} orchestration core manages layer ordering, scheduling, exchange, and feedback and coordinates the functions of the third, fourth, fifth, and seventh layers; however, it does not replace the models or controller. Dynamic management of relationships remains theoretical.

\subsection{Comparison of Related Approaches and Positioning of the Present Study}

MINS\footnote{Multisensor-Aided Inertial Navigation System} aligns and calibrates asynchronous data without evaluating trans-domain control \cite{ref04_lee2025mins}. Integrity-aware navigation connects health monitoring and source recovery to estimation but depends on cameras and lidar \cite{ref06_aljaafreh2026integrity}. The focus of the present research is joint decision-making across navigation, dynamics, and environment.

In control-oriented twins, Yang et al. used MPC\footnote{Model Predictive Control} and feedback for obstacle avoidance \cite{ref23_yang2025digital}; Kumar et al. connected a synchronized model and visual-reference-based indoor navigation to command execution \cite{ref24_kumar2026indoor}. Explicit coupling among multiple specialized twins was not reported; this does not imply its absence.

Gil et al. investigated the co-simulation of heterogeneous models \cite{ref08_gil2026model}; OpenTwins studied semantic networking and runtime adaptation \cite{ref25_robles2026composite}. Applied TDDT, involving coordination of specialized models and online adaptation, has been evaluated in a journal study \cite{ref22_amiri2026greenhouse}; the TDDT formalism defines shared state, six types of coupling, TDOC, and multirate coordination \cite{ref09_amiri2026transdomain}. Dynamic N-Twin is proposed and has not been fully implemented. Here, Native denotes the Linux-based integration of TDDT–Phoenix with ArduPilot SITL\footnote{SITL (Software-in-the-Loop): A method for testing and validating a control system by executing the actual software in interaction with a simulator of physical behavior, without requiring physical hardware.} and JSBSim, not hardware testing or physical flight.

As summarized in Table~\ref{tab:related_approaches}, TDDT--Phoenix links three specialized twins through prediction, command execution, and traceable feedback. Nevertheless, the evidence is limited to simulation and does not establish statistical superiority or full realization of Dynamic N-Twin.

\begin{table*}[htbp]


\definecolor{TDDTNavy}{HTML}{24364D}
\definecolor{TDDTStripe}{HTML}{F0F4F9}
\definecolor{TDDTGreen}{HTML}{DDF2E9}
\definecolor{TDDTYellow}{HTML}{FFF0D8}
\definecolor{TDDTGray}{HTML}{E7ECF2}
\definecolor{TDDTRed}{HTML}{F7E1E5}
\definecolor{TDDTText}{HTML}{24364D}

\centering

\caption{Comparison of Related Approaches}

\label{tab:related_approaches}

\vspace{6pt}

\begingroup

\footnotesize
\color{TDDTText}

\setlength{\tabcolsep}{4pt}
\renewcommand{\arraystretch}{1.3}

\begin{tabularx}{\textwidth}{
    >{\raggedright\arraybackslash}p{0.18\textwidth}
    >{\raggedright\arraybackslash}X
    >{\raggedright\arraybackslash}X
    >{\raggedright\arraybackslash}X
    >{\raggedright\arraybackslash}X
}

\toprule


\rowcolor{TDDTNavy}

\textcolor{white}{\textbf{Approach / Reference}} &

\textcolor{white}{\textbf{Integration and Domain Contributions}} &

\textcolor{white}{\textbf{Decision-Making and Closed-Loop Operation}} &

\textcolor{white}{\textbf{Online Adaptation and Uncertainty}} &

\textcolor{white}{\textbf{Validation}} \\

\midrule


\rowcolor{TDDTStripe}

MINS \cite{ref04_lee2025mins} &

Fusion of navigation sources; temporal alignment &

Joint estimation; control not reported &

Calibration and covariance &

Simulation and real-world data \\

\addlinespace


Integrity-aware navigation
\cite{ref06_aljaafreh2026integrity} &

Navigation sources with health monitoring &

Source switching and recovery; associated control &

Fault detection and covariance management &

Flight tests and replay \\

\addlinespace


\rowcolor{TDDTStripe}

Flight-control twin
\cite{ref23_yang2025digital} &

Aircraft, trajectory, and control models &

MPC and physical feedback &

Comprehensive cross-domain adaptation not reported &

Physical aircraft test \\

\addlinespace


Indoor-navigation twin
\cite{ref24_kumar2026indoor} &

Vision, trajectory, and synchronized model &

Commands and closed-loop feedback &

Statistical integrity not reported &

Marker-based indoor experiment \\

\addlinespace


\rowcolor{TDDTStripe}

Robotic co-simulation
\cite{ref08_gil2026model} &

Heterogeneous models; execution coordination &

Joint decisions depend on the application &

Uncertainty management depends on the model &

Case study \\

\addlinespace


OpenTwins
\cite{ref25_robles2026composite} &

Semantic network of twins &

Event-driven interaction; physical control not reported &

Runtime adaptation &

Architecture and scalability evaluation \\

\addlinespace


\rowcolor{TDDTStripe}

Applied TDDT
\cite{ref22_amiri2026greenhouse} &

Specialized models and aligned state &

Joint decision-making; application-level loop &

Model-based adaptation &

Journal case evaluation \\

\addlinespace


TDDT and Dynamic N-Twin
\cite{ref09_amiri2026transdomain} &

Shared state; six types of coupling &

Trans-domain decision-making; multirate loops &

Dynamic trust and topology; proposed &

Theoretical architecture \\

\addlinespace


\rowcolor{TDDTStripe}

Present study (TDDT--Phoenix) &

Three specialized twins; atomic state &

Limited prediction, executable commands, and model feedback &

Limited correction; full network not active &

Native simulation test with independent evaluation \\

\bottomrule

\end{tabularx}

\endgroup

\end{table*}

\section{Proposed Architecture of the Trans-Domain Digital Twin System}

This section describes the responsibilities of the layers, information exchange, and the command and feedback cycle in TDDT--Phoenix.

\subsection{Seven-Layer System Architecture}

In the seven-layer TDDT--Phoenix architecture \cite{ref09_amiri2026transdomain}, the first layer acquires sensor data, actuator states, and physical constraints from the UAV and its environment. In the second layer, MESA-NAV estimates the motion state, WingMPET computes the dynamic response, and MicroMet models atmospheric conditions. Each twin reports information quality and uncertainty along with its outputs.

The third layer aligns the outputs in terms of time, semantics, and coordinate frames and registers them in the Shared Trans-Domain State (STD). In the Native implementation, an atomic snapshot contains the states of the three twins, active relationships, constraints, mission reference, time, and an integrity identifier. The fourth layer predicts the consequences and feasibility of candidate control actions, while the fifth layer generates constrained and traceable commands through limited predictive selection and safety checks. However, comprehensive risk and uncertainty management has not been implemented.

After verifying command validity, the sixth layer transmits the command to ArduPilot through the digital pilot and graphical interface, while JSBSim simulates the flight response. The seventh layer compares the prediction associated with the same transaction against sensor-based observations obtained after execution and enables only limited correction of WingMPET. The simulator's ground-truth position serves exclusively as an independent evaluation reference.

\begin{figure}[htbp]
    \centering
    \includegraphics[width=0.6\linewidth,alt={Seven-layer TDDT–Phoenix architecture integrating MESA-NAV, WingMPET, and MicroMet through shared-state coordination, predictive decision-making, simulated command execution, and sensor-based feedback. The implemented fixed coupling pathway is distinguished from the proposed Dynamic N-Twin extension.}]{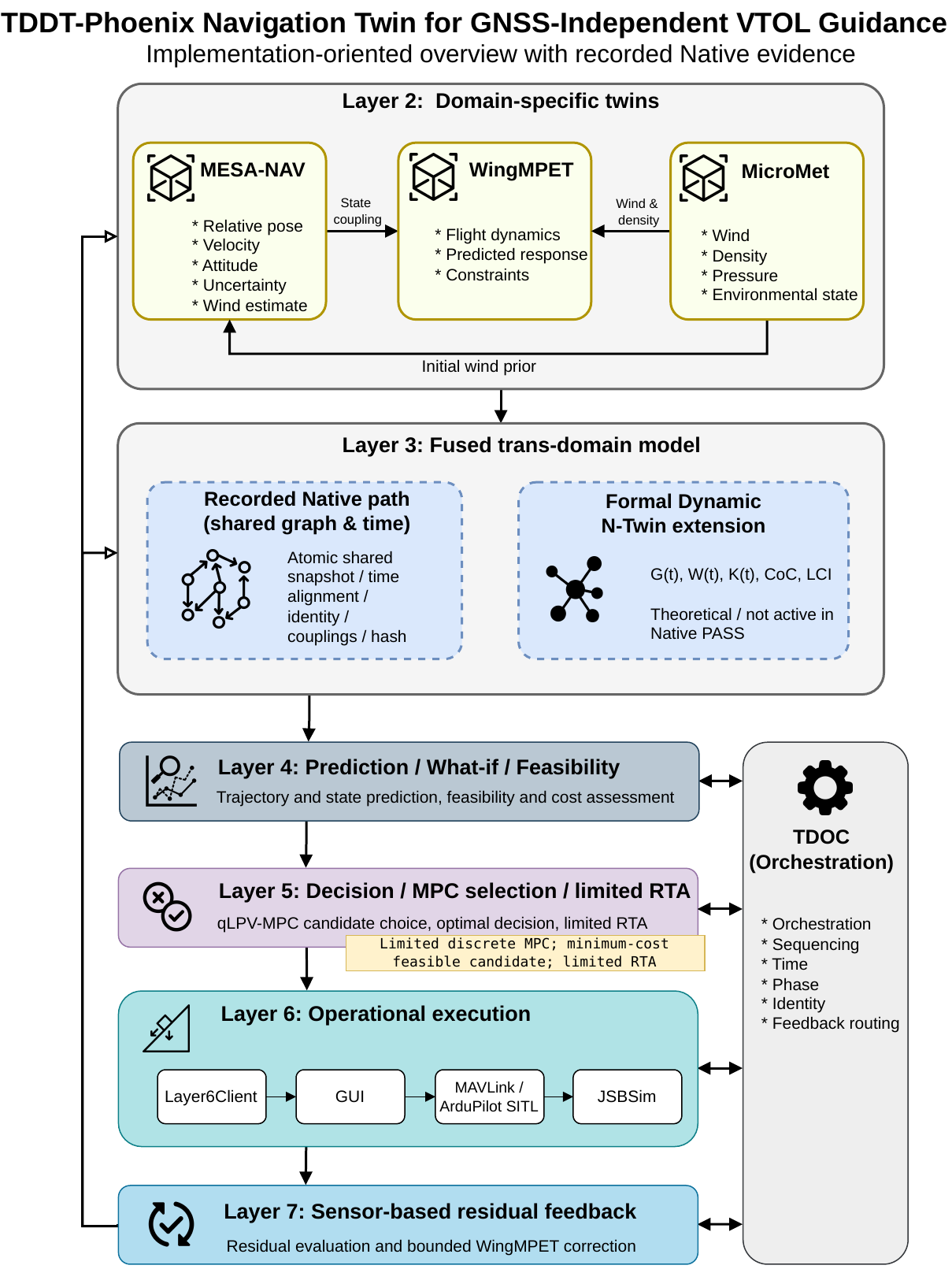}
    \caption{Implementation-oriented architecture of TDDT–Phoenix for GNSS-independent VTOL guidance. Layer 2 comprises the MESA-NAV, WingMPET, and MicroMet twins, whose state and environmental exchanges are integrated through atomic, time-aligned snapshots in Layer 3. The recorded Native coupling pathway is distinguished from the proposed Dynamic N-Twin extension, including topology G(t), epistemic weights W(t), coupling K(t), and consistency indicators (CoC/LCI), which were not fully implemented or validated. In the Native implementation, Layers 4 and 5 perform limited receding-horizon prediction and constrained selection among at most nine phase-dependent control candidates; full qLPV-MPC optimization and global optimality were not demonstrated. Layer 6 provides the command pathway through ArduPilot SITL and JSBSim. Layer 7 supports sensor-based residual feedback for bounded WingMPET correction, while TDOC coordinates sequencing, timing, mission phases, transaction identity, and feedback routing.}
    \label{fig:tddt_phoenix_architecture}
\end{figure}

TDOC coordinates layer sequencing, mission phases, timing, and identifier continuity. Fast, intermediate-scale, and slow loops are defined in the formalism; however, the dynamic topology of Dynamic N-Twin and comprehensive cross-domain error correction have not been validated in the Native implementation. The implementation-oriented architecture of TDDT–Phoenix, including inter-twin coupling, shared-state integration, prediction, decision-making, operational execution, and sensor-based feedback under TDOC coordination, is illustrated in Figure \ref{fig:tddt_phoenix_architecture}.

Quasi-Linear Parameter-Varying Model Predictive Control (qLPV-MPC) is a predictive control approach that approximates nonlinear system dynamics using parameter-dependent linear models to optimize control actions over a finite prediction horizon while satisfying operational constraints.

Center of Consistency (CoC) is a shared reference state that minimizes weighted discrepancies among interconnected digital twins while satisfying cross-domain consistency constraints. Also, the Loop Current Index (LCI) measures the deviation between a digital twin’s observed state and its direct prediction, indicating the influence of feedback loops within the interconnected twin network.

\subsection{Domain-Specific Digital Twins}

MESA-NAV uses gyroscope, accelerometer, magnetometer, barometer, pitot sensor, time, and environmental priors to estimate relative position, velocity, orientation, and wind. A three-dimensional anemometer was designed but did not provide operational data in the Native implementation. Heading and altitude corrections were implemented, whereas full multi-field anchoring and comprehensive covariance estimation have not been validated. Navigation outputs and limited uncertainty information are transferred to WingMPET and the third layer.

WingMPET receives the MESA-NAV state, Phoenix mass and geometry, actuator outputs, wind, air density, elevation maps, and constraints. Its formalism covers the six-degree-of-freedom dynamic response, path, time, energy, safety clearance, and model validity. For short-term prediction, the Native implementation uses the current mechanical response, thrust capacity, and initial acceleration. However, comprehensive path--time--energy optimization and the complete transition to fixed-wing flight have not been validated.

MicroMet converts meteorological data, requested location and time, and a digital elevation model into local fields of temperature, humidity, pressure, and wind. Precipitation, radiation, and air density are incorporated into its outputs or supplementary relationships. The MicroMet-3DW\footnote{MicroMet Three-Dimensional Wind (MicroMet-3DW) extension} extension diagnostically estimates vertical wind, gradients, and temporal variations, but it is neither a complete turbulence model nor an independent measurement system. In the Native implementation, time-stamped wind and air density data were delivered to WingMPET. Their validity depends on the initial data, spatial resolution, terrain characteristics, and calibration.

The implemented coupling is asymmetric: MESA-NAV provides motion states, while MicroMet supplies wind and air density to WingMPET. Estimated position and time determine the MicroMet requests. However, MESA-NAV receives only the initial wind prior, and its position is not continuously corrected. Feedback provides only limited correction of the WingMPET disturbance parameter.

\subsection{Shared State and Cross-Domain Interconnections}

The third layer constructs STD from the domain-specific states of the three twins, mission constraints, and metadata describing quality, uncertainty, and provenance. Each twin remains responsible for its own model computations. Measurement, generation, cycle, and consumption times are recorded separately. Data freshness is assessed according to source timestamps and validity conditions. Limited interpolation or propagation is permitted, but expired data cannot regain validity without fresh evidence.

Alignment of units, coordinate frames, and variable semantics preserves the distinction between ground velocity, airspeed, and the wind vector. In the Native implementation, an atomic snapshot controlled for temporal and identity consistency fixes the states of the three twins, static directed relationships, constraints, mission reference, and transaction identifier. The snapshot identifier and integrity marker are preserved throughout prediction, decision-making, and command generation to prevent outputs from different cycles from being mixed.

\begin{figure}[htbp]
    \centering
    \includegraphics[width=0.90\linewidth,alt={Three-part Layer 3 architecture showing shared-state alignment, the proposed Dynamic N-Twin relationship network, and multirate temporal coordination. Arrows connect specialized twin outputs, mission constraints, TDOC orchestration, atomic snapshots, and prediction-layer inputs.}]{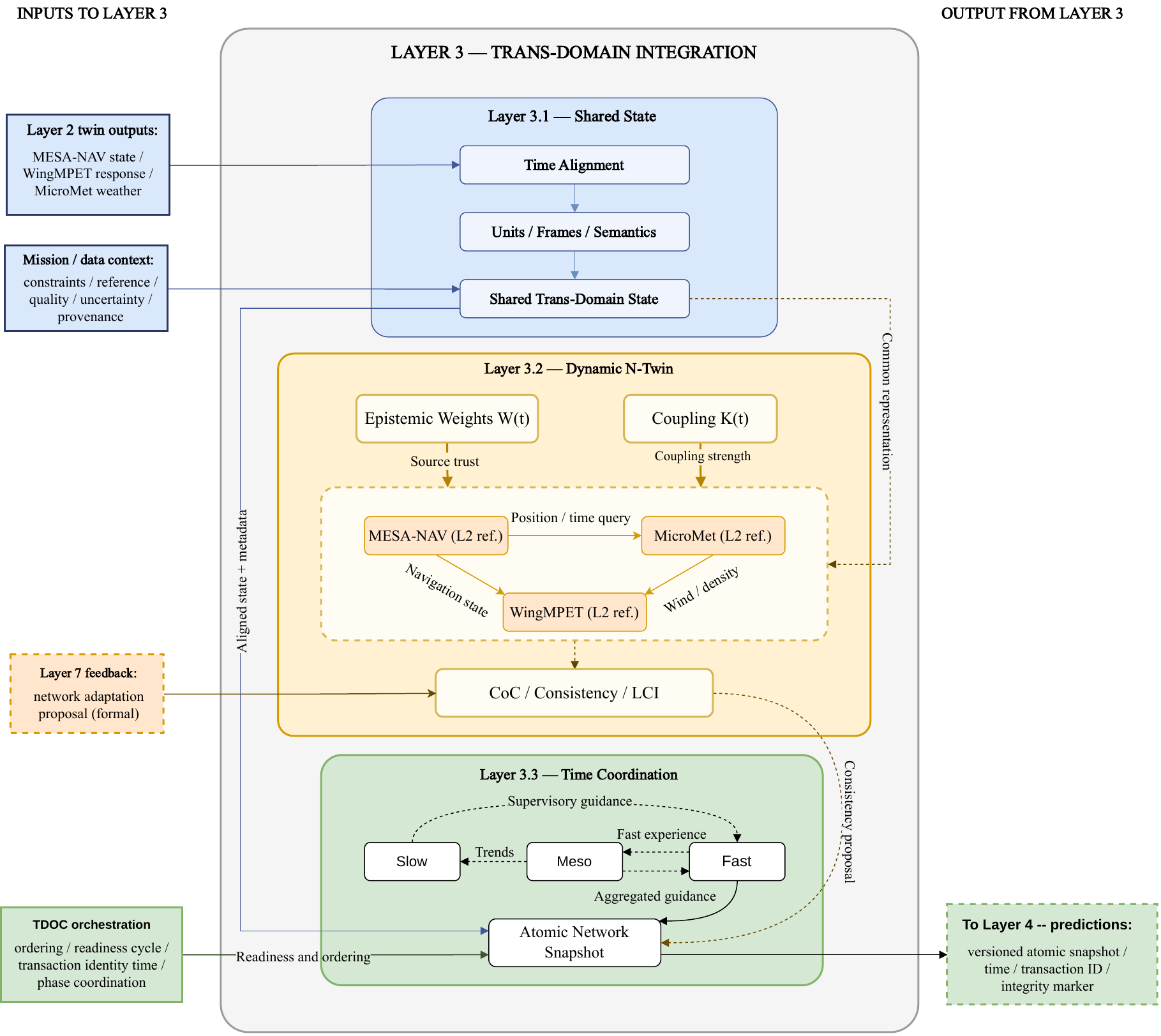}
    \caption{Conceptual organization of Layer 3 and its limited Native realization. Sublayer 3.1 represents alignment, provenance checking, and construction of an atomic shared-state packet from heterogeneous twin outputs. The recorded Native packet should not be interpreted as a complete implementation of the formal Shared Trans-Domain State. Sublayer 3.2 depicts the proposed Dynamic N-Twin mechanisms, including topology G(t), epistemic weights W(t), coupling strengths K(t), and consistency indicators (CoC/LCI); the evaluated Native implementation recorded fixed declared relationships rather than these complete dynamic mechanisms. Sublayer 3.3 illustrates temporal coordination and atomic snapshot delivery. Full dynamic-network and multirate formalism capabilities were not independently validated in the Native tests.}
    \label{fig:tddt_layer3_node_refs_layout_diagram}
\end{figure}

The implemented couplings include transferring navigation states to the dynamics model, atmospheric conditions to the dynamics model, and requesting atmospheric data based on estimated position and time. The complete formalism specifies six types of coupling: data, model, state, error, objective, and control. Late, incomplete, or inconsistent data are restricted or rejected. However, dynamic trust weighting, the consistency center, and comprehensive assessment of physical inconsistencies have not been implemented. TDOC orchestrates the delivery of valid snapshots to the prediction and decision-making processes, while valid feedback enables only limited correction of WingMPET in the subsequent cycle. Figure \ref{fig:tddt_layer3_node_refs_layout_diagram} illustrates the internal organization of Layer 3, comprising shared-state construction (3.1), the conceptual network of inter-twin coupling relationships and their attributes (3.2), and multirate temporal coordination with atomic snapshot generation (3.3).

\subsection{Trans-Domain Orchestration Core}

TDOC manages the sequence of component invocations and the flow of information \cite{ref09_amiri2026transdomain}, while domain-specific computations and command selection remain outside its responsibilities. Data flow, invocation, delivery, and feedback are treated as distinct processes. Progression to the next stage depends on component readiness, operational health, timing constraints, and version and data consistency. During each cycle, TDOC coordinates updates to the three twins. The third-layer snapshot is then passed to the fourth layer for prediction, the fifth for decision-making, the sixth for execution, and the seventh for feedback. This causal sequence does not require simultaneous execution.

In the formalism, the fast loop handles the current state and decisions, the intermediate-scale loop aggregates evidence and provides phase-specific guidance, and the slow loop revises policies and constraints \cite{ref09_amiri2026transdomain}. TDOC distinguishes measurement, simulation, prediction, decision, transmission, feedback, and wall-clock times. State and policy versions remain fixed within each transaction, while the safe commitment of changes at synchronization boundaries is defined in the complete design.

The mission begins with readiness and calibration checks before ARM\footnote[5]{ARM (arming) refers to the readiness of the propulsion system to receive commands and rotate the motors, but does not necessarily indicate that the motors are rotating or that flight has begun.}. It includes arming confirmation, vertical takeoff, VTOL movement, descent, landing flare, contact stabilization, and confirmed DISARM. Phase transitions depend on altitude, velocity, and evidence of contact. Estimated proximity to the destination does not establish mission success or physical contact. The complete transition to fixed-wing flight has not been tested.

The exchange contract specifies data semantics, units, coordinates, source and destination, test and transaction identities, validity time, version, quality, and provenance. In addition to decision acceptance, the sixth layer checks the mission phase, data freshness, and execution readiness. Decision generation, command acceptance, transmission, receipt acknowledgment, and physical response are distinct stages. TDOC matches command and feedback identities, operating mode, ARM state, and temporal sequence. Late or mismatched observations are not attributed to a new cycle. Stopping and degraded operation are included in the formalism, but comprehensive management of topology, trust, and version recovery has not been validated in the Native implementation.

\subsection{Offline Knowledge and Online Adaptation}

Offline knowledge is derived from scenarios and historical evidence as model and mission priors after assessing quality, coverage, provenance, and version information \cite{ref09_amiri2026transdomain}. This knowledge does not replace live observations or authorization to execute commands, and its use depends on model compatibility and the transfer contract.

CRP\footnote{Contextual Reference Patterns}\cite{ref09_amiri2026transdomain} partition environmental and mission states into matchable clusters. SARG\footnote{Stage-Aware Reference Guidance}\cite{ref09_amiri2026transdomain} proposes reference actions, performance, constraints, and objective weights according to the flight regime, mission phase, and CRP. Its acceptance depends on coverage and confidence. TEP\footnote{Twin Epistemic Profile}\cite{ref09_amiri2026transdomain} distinguishes model quality, uncertainty, and validity range. In addition, SS-KStore\footnote{Step-Stream Memory Knowledge Store}\cite{ref09_amiri2026transdomain} retains experience from fast cycles, while SETD-KStore\footnote{Single-Episode Trans-Domain Knowledge Store}\cite{ref09_amiri2026transdomain} stores condensed mission experience. In the formalism, Mesoscale Reference Guidance (MRG) can retrieve CRP/SARG information to adjust decision constraints and weights for the subsequent cycle.

In the Native implementation, CRP, SARG, and TEP provided contextual information, while current decisions relied on fresh data. The only validated adaptation was the limited correction of WingMPET disturbances using valid feedback. Continuous learning and complete dynamic adaptation of the network were not achieved.

\section{Implementation Details}

This section describes the active algorithms, execution constraints, and software integration of TDDT--Phoenix in the Native test.

\subsection{Sensor Data Processing and State Estimation}

Without online GNSS, cameras, or radar, MESA-NAV estimates motion using a three-axis IMU\footnote{Inertial Measurement Unit}, heading using a magnetometer, relative altitude using a barometer, and airspeed magnitude using a pitot sensor. Inertial, magnetic, and pressure data are obtained from \texttt{HIGHRES\_IMU}, while airspeed is obtained from \texttt{VFR\_HUD}. The initial references comprise a known origin, a reference magnetic field, initial wind from MicroMet, and prior time information. The three-dimensional anemometer specified in the formalism was not active in Native.

The requested IMU and airspeed sampling rates are \(100\,\mathrm{Hz}\) and \(20\,\mathrm{Hz}\), respectively, while the TDOC target frequency is \(10\,\mathrm{Hz}\). Each IMU sample is processed after checking its identity, source timestamp, ordering, freshness, and coordinate frame. The maximum permitted ages of IMU and pitot samples are \(0.3\,\mathrm{s}\) and \(0.5\,\mathrm{s}\), respectively, and the maximum propagation step is \(0.15\,\mathrm{s}\). Late or duplicate samples do not trigger independent corrections; software alignment does not constitute physical synchronization.

A \(60\,\mathrm{s}\) stationary calibration before ARM establishes the initial inertial biases, barometric reference, and orientation. Initial orientation is obtained using Wahba's method based on gravity and the magnetic field. The stationary constraint is released during takeoff, and units and NED\footnote{North-East-Down}/FRD\footnote{Forward-Right-Down} coordinate frames are made consistent. The magnetic reference depends on spatial tables and estimated position; however, heading correction does not constitute magnetic localization, and complete compensation for motor interference has not been verified. Barometric altitude is corrected only after a valid update and an NIS\footnote{Normalized Innovation Squared} acceptance test.

After bias removal and quaternion propagation, relative position at each step is calculated using a constant-acceleration approximation:

\[
\hat{\mathbf p}_{k+1}^{n}
=
\hat{\mathbf p}_{k}^{n}
+
\hat{\mathbf v}_{k}^{n}\Delta t
+
\tfrac12
\left[
\mathbf R_b^n
\left(
\mathbf f_{m,k}-\hat{\mathbf b}_{f,k}
\right)
+
\mathbf g^n
\right]
\Delta t^2.
\]

In this equation, \(\hat{\mathbf p}^{n}\) denotes relative position (meters), \(\hat{\mathbf v}^{n}\) ground velocity (meters per second), \(\mathbf R_b^n\) the body-to-navigation transformation, \(\mathbf f_m\) measured specific force, \(\hat{\mathbf b}_f\) its bias, and \(\mathbf g^n\) gravity (all three expressed in meters per second squared). Furthermore, \(\Delta t\) denotes the time step (seconds), and \(k\) is the sample index. The active model maintains a complementary \(9\times9\) covariance for position, velocity, and wind, rather than the 29-dimensional error-state filter defined in the formalism.

The pitot sensor constrains the magnitude of relative airspeed but cannot determine ground velocity without independent wind information. Invalid observations are rejected, although a contaminated state is not automatically recovered. The outputs include relative position and velocity, orientation, altitude, estimated wind, and complementary covariance. Bias, pressure, magnetic interference, and wind--velocity ambiguity cause drift, while absolute horizontal position remains unobservable without a spatial anchor. Neither \texttt{SIM\_STATE} nor the controller's precomputed state estimates are injected into MESA-NAV. Timestamped outputs with provenance information are transferred to Layer 3; multi-field anchoring and comprehensive confidence estimation have not been implemented.

\subsection{Flight Dynamics and Atmospheric Conditions Modeling}

WingMPET estimates the motion response of Phoenix using the MESA-NAV state, actuators, and MicroMet. The formalism defines position, velocity, quaternion, body rates, motor rotational speeds, and battery charge within a six-degree-of-freedom model. In Native, only one physical-model step is used, followed by a reduced-order model of the internal ArduPilot controller. The full simulation horizon is not executed using the complete physical model, and JSBSim is not replaced.

The active model accounts for a mass of \(0.65\,\mathrm{kg}\), two motors, two elevons, thrust saturation, and propeller wash. Motor rotational speeds and elevon angles are reconstructed from servo outputs. The propeller-wash relationships are:

\[
\left(
L_{pw,i},
D_{pw,i},
M_{pw,i}
\right)
=
\omega_i^2
\left(
k_L\delta_i,\;
k_D^{pw}\delta_i^2,\;
k_M^{pw}\delta_i
\right).
\]

In this equation, \(L,D\) denote forces (newtons), \(M\) denotes moment (newton-meters), \(\omega_i\) is the rotational speed of motor \(i\) (radians per second), and \(\delta_i\) is the elevon deflection (radians). Assuming radians are dimensionless, \(k_L,k_D^{pw}\) are force coefficients (newton-seconds squared), and \(k_M^{pw}\) is a moment coefficient (newton-meter-seconds squared). The smooth hover-to-fixed-wing transition depends on positive chordwise velocity; however, the fixed-wing coefficients have low confidence, and the complete transition has not been tested. Initial acceleration is obtained from the velocity change over a \(0.01\,\mathrm{s}\) WingMPET step.

MicroMet uses timestamped atmospheric data, estimated position, and a digital elevation model (DEM)\footnote[5]{\textbf{DEM (Digital Elevation Model):} A digital elevation model represents the elevations of points on the Earth's surface relative to a specified vertical reference.} to interpolate and apply terrain-related corrections to temperature, humidity, pressure, and wind. Air density is calculated from thermodynamic relationships, while vertical wind is obtained from the diagnostic 3DW extension. The model is invoked approximately every \(0.5\,\mathrm{s}\), and data freshness, provenance, and DEM validity are checked. However, it is neither a CFD\footnote{Computational Fluid Dynamics} solver nor a validated gust-prediction model. The air velocity supplied to the dynamics model is calculated as:

\[
\mathbf v_a^b
=
\mathbf R_n^b
\left(
\mathbf v_g^n-\mathbf w^n
\right).
\]

In this equation, \(\mathbf v_a^b\) is body-frame air velocity, \(\mathbf v_g^n\) is ground velocity, \(\mathbf w^n\) is wind in the NED\footnote[6]{\textbf{NED (North--East--Down):} A three-axis local coordinate system oriented toward north, east, and down, used to represent UAV position, velocity, and motion.} frame (all expressed in meters per second), and \(\mathbf R_n^b\) is the navigation-to-body transformation. Wind and air density affect forces, acceleration, trajectory, and time. However, the Environment Adapter sends only supported variables with evidence of application to JSBSim, rather than necessarily applying all MicroMet outputs.

The active Layer 4 screening evaluates acceleration, thrust reserve, airspeed, body rates, and candidate validity. However, comprehensive collision, terrain-referenced altitude, energy, and transition constraints; trajectory--time--energy optimization; and actual battery consumption have not been validated. Prediction validity depends on state estimation, aerodynamics, and atmospheric conditions.

\subsection{Prediction and Optimization of Control Decisions}

Layer 4 receives an atomic snapshot containing the MESA-NAV state, WingMPET response, MicroMet atmospheric conditions, mission phase, and mission reference. Reference velocity is calculated as \(0.45\) times the position error, subject to a phase-dependent limit, while the desired horizontal acceleration is obtained as \(0.25\) times the velocity error minus the disturbance. Offline knowledge provided only the model identity for the current decision.

The acceleration variation along each horizontal axis is selected from the set \(0\) and \(\pm0.1\,\mathrm{m/s^2}\), producing up to nine control candidates. The candidate set is restricted by mission phase: the recorded third test contained 4,323 cycles with nine candidates and 132 cycles with a single phase-permitted candidate during the final landing and contact-stabilization phases. Initial acceleration is obtained from one WingMPET step, while the remainder of the horizon is calculated using the reduced-order first-order response of ArduPilot:

\[
\mathbf a_{c,\ell+1}
=
\mathbf a_{c,\ell}
+
\left(
1-e^{-\Delta t/\tau}
\right)
\left(
\mathbf a_c-\mathbf a_{c,\ell}
\right).
\]

In this equation, \(\mathbf a_{c,\ell}\) denotes the predicted acceleration of candidate \(c\) at step \(\ell\), \(\mathbf a_c\) is the target acceleration (meters per second squared), \(\Delta t=0.1\,\mathrm{s}\), and \(\tau=1\,\mathrm{s}\) is the response time constant. The two-second horizon contains 20 steps; neither the complete six-degree-of-freedom model nor JSBSim is executed during the rollout.

After checking the hash, transaction, values, and feasibility, Layer 5 selects the lowest-cost candidate:

\[
c_k^*
=
\arg\min_{c\in\mathcal C_k^{\mathrm{feasible}}}
\left(
J_{v,c}
+
0.15J_{p,c}
+
0.025J_{a,c}
+
0.5J_{\Delta a,c}
\right).
\]

In this equation, \(c_k^*\) is the selected candidate from the feasible set \(\mathcal C_k^{\mathrm{feasible}}\). The four \(J\) terms respectively represent the costs of velocity error, position error, acceleration magnitude, and acceleration variation relative to the previous command; these are soft criteria. Reselection at each cycle constitutes discrete and constrained MPC, rather than continuous or global optimization. The command is generated through a pilot-control mapping based on the selected candidate and estimated state.

Initial screening checks a maximum commanded lean angle of \(8^\circ\), thrust reserve with an \(18^\circ\) reference, and finiteness of values. The reference horizontal speed is limited to \(6\,\mathrm{m/s}\), while maximum climb and descent rates are \(1.2\,\mathrm{m/s}\) and \(0.8\,\mathrm{m/s}\), respectively. The separate RTA\footnote{Runtime Assurance} gate monitors observed airspeed up to \(18\,\mathrm{m/s}\), lean angle up to \(25^\circ\), and Roll/Pitch/Yaw rates up to \(300/200/150\,\mathrm{deg/s}\), respectively. During Flare, only leveling commands are permitted; after valid contact, only \texttt{CONTACT-SUPPORT} in \texttt{QSTABILIZE} with zero throttle is allowed.

TDOC targets a decision frequency of \(10\,\mathrm{Hz}\), while the command service operates at \(30\,\mathrm{Hz}\). Prediction starts from the snapshot source timestamp, and only the first command is executed. Command validity is \(0.5\,\mathrm{s}\), and the maximum state age at transmission is \(0.3\,\mathrm{s}\). The maximum recorded cycle durations in the first, second, and third full-route tests were 153.35, 171.63, and \(181.71\,\mathrm{ms}\), respectively (Section 6.3); therefore, strict compliance with the nominal \(100\,\mathrm{ms}\) cycle deadline was not demonstrated. The absence of a feasible candidate, valid mode, or consistent snapshot causes normal commands to be rejected, and stale packets are not silently substituted. Comprehensive covariance propagation, Tube MPC, chance constraints, Pareto optimization, collision/energy constraints, and guarantees of stability or recursive feasibility have not been validated.

\subsection{Flight Command Execution and Direct System Integration}

Layer 5 sends the pilot-control request \(u_{\mathrm{cmd}}=[\mathrm{aileron},\mathrm{elevator},\mathrm{throttle},\mathrm{rudder}]\), together with the mode, phase, timestamp, deadline, transaction/candidate identifiers, and \texttt{snapshot\_hash}. This request is neither a PWM\footnote[10]{\textbf{PWM (Pulse Width Modulation):} Pulse-width modulation is a method of controlling actuators by varying the width of command-signal pulses. In the Phoenix UAV model, four PWM-equivalent command channels are available: two channels control the rotational speeds of the left and right motors, and two control the deflection angles of the left and right elevon control surfaces.} command nor a direct thrust command. Acceleration is transformed into the tail-sitter frame using \(\mathrm{Ry}(-90^\circ)\) and the NED-to-Forward/Right transformation. Roll/Pitch commands are derived from acceleration and gravity, while command limits are determined by \texttt{Q\_A\_ANGLE\_MAX} and the deadzone. In \texttt{QHOVER}, throttle requests the climb rate.

\begin{figure}[htbp]
\centering
\includegraphics[width=0.96\linewidth,alt={Sequential command-execution pathway from predictive decisions and safety validation through Layer6Client, command intake, semantic latch, Digital Pilot, MAVLink, ArduPilot SITL, and JSBSim, followed by sensor-based feedback to Layer 7.}]{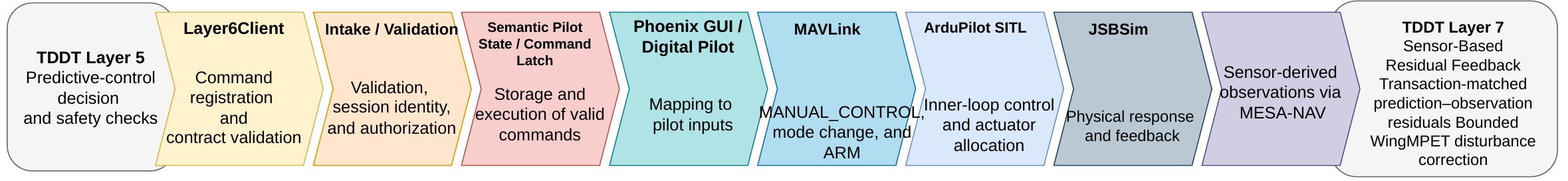}
\caption{Command-execution pathway in the Native TDDT–Phoenix implementation. Predictive control decisions and safety checks are transferred through Layer6Client, intake validation, the semantic command latch, and Digital Pilot to ArduPilot SITL via MAVLink. ArduPilot manages actuator allocation, while JSBSim simulates the resulting physical response. }
\label{fig:layer6_details_command-execution_pathway}
\end{figure}

Layer 6 performs \texttt{intake/validation} to check identity, transaction, version, freshness, mode, phase, and values. The permitted ranges for Aileron/Elevator/Rudder and Throttle are \([-1,1]\) and \([0,1]\), respectively. Intake acceptance indicates only that the command has been recorded in the \texttt{command latch} and pilot state; it does not confirm transmission or response. Invalid commands are not executed. Comprehensive actuator health monitoring and atomic multi-actuator execution remain theoretical capabilities. The implemented command-execution pathway, from predictive decision-making and Layer 6 validation to Digital Pilot, MAVLink\footnote{Micro Air Vehicle Link} transmission, ArduPilot SITL, and the simulated physical response in JSBSim, is illustrated in Figure \ref{fig:layer6_details_command-execution_pathway}.

\begin{figure}[htbp]
\centering
\includegraphics[width=0.55\linewidth,alt={Digital Pilot graphical interface displaying virtual joystick controls, directional inputs, arming and flight-mode options, and real-time telemetry and status fields for Phoenix VTOL simulation.}]{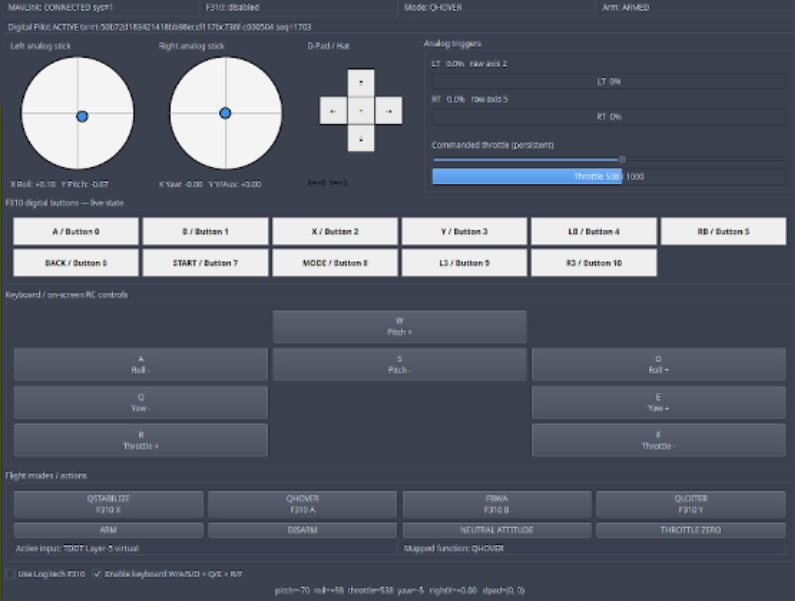}
\caption{Digital Pilot graphical user interface used in the Native ArduPilot/JSBSim integration. The interface provides pilot-control inputs, operating-mode selection, and telemetry and status monitoring. Control requests are transmitted through the MAVLink interface, while the flight controller and simulator determine the resulting vehicle response.}
\label{fig:phoniex_gui}
\end{figure}

In Digital Pilot, the GUI\footnote{Graphical User Interface} (Fig. \ref{fig:phoniex_gui}) sends \texttt{MANUAL\_CONTROL} and \texttt{MODE} and \texttt{ARM/DISARM} requests through MAVLink 2 without simultaneous gamepad use. RealtimeIO transfers telemetry and parameters. The test includes ArduPilot SITL, JSBSim, the Environment Adapter, GUI, and MAVProxy Map/Console, with heartbeat, identity, and freshness checks. ArduPilot converts commands into actions for the two motors and two elevons according to the active mode, while JSBSim simulates a response independently of WingMPET. \texttt{SERVO\_OUTPUT\_RAW} supplies actuator reconstruction, whereas \texttt{SIM\_STATE} is used only for visualization/validation. Virtual GNSS is disabled. \texttt{QHOVER/QSTABILIZE} require valid Pre-Arm conditions and state information; absolute horizontal position is not mandatory.

The transmission frequency is \(30\,\mathrm{Hz}\), command validity is \(0.5\,\mathrm{s}\), the state-age limit is \(0.3\,\mathrm{s}\), and the Receipt deadline is \(0.12\,\mathrm{s}\). The two-second horizon does not constitute an execution commitment. Source, transmission, and observation timestamps are recorded separately. After calibration and Pre-Arm checks, ARM is confirmed through Heartbeat. The \texttt{vertical\_takeoff}, \texttt{cruise\_vtol}, \texttt{vertical\_landing}, \texttt{landing\_flare}, and \texttt{contact\_settle} phases are then executed. During Flare, horizontal/Yaw commands are set to zero, and descent is constrained. Barometer/IMU-based contact evidence is required for \texttt{CONTACT-SUPPORT} with zero throttle. Mission completion requires a confirmed \texttt{landed\_state}, stability, and DISARM.

Receipt matches test/transaction/command identifiers, mode, command vector, ordering, \texttt{MANUAL\_CONTROL/TIMESYNC}, and MAVLink transmission. Discontinuity, expiration, or lack of response invalidates the evidence. Layer 7 accepts feedback associated with the same transaction only after valid transmission and sensor-based observation. Integration has been verified in SITL; hardware execution, fixed-wing transition, and comprehensive fault tolerance have not been validated.

\begin{figure}[htbp]
\centering
\includegraphics[width=0.60\linewidth,alt={Satellite-based terrain map with a yellow spatial grid, displaying the simulated Phoenix VTOL position, a home-location marker, surrounding geographic features, and flight-monitoring status information.}]{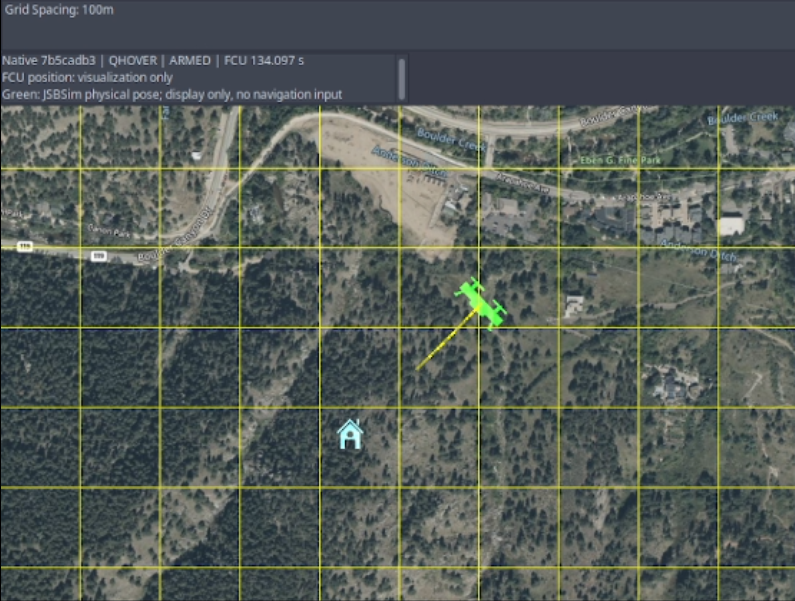}
\caption{Geographic visualization of the Phoenix VTOL during Native software-in-the-loop execution. The map displays the simulated vehicle position and surrounding terrain for operational monitoring.}
\label{fig:phoniex_map}
\end{figure}

\begin{figure}[htbp]
\centering
\includegraphics[width=0.70\linewidth,alt={MAVProxy Console screenshot showing the Phoenix UAV in QHOVER mode, flight telemetry, vehicle status, and messages confirming acceptance of the ARM command during software-in-the-loop simulation.}]{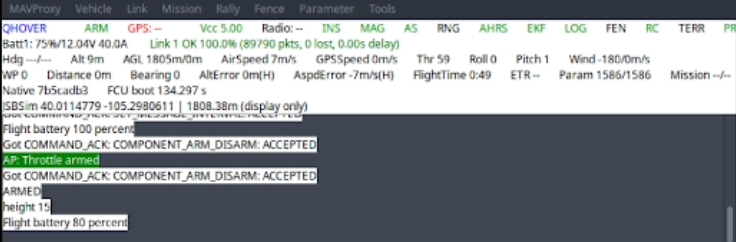}
\caption{MAVProxy Console snapshot from the Native software-in-the-loop environment, displaying the QHOVER flight mode, vehicle telemetry, and ARM command acknowledgment.}
\label{fig:phoniex_console}
\end{figure}

\subsection{Error Feedback and Model Adaptation}

Layer 7 matches the candidate prediction, selected command, transmission Receipt, and post-execution observation from MESA-NAV using the transaction identifier, \texttt{snapshot\_hash}, candidate, command sample, and source timestamp. Ground-truth \texttt{SIM\_STATE}, GNSS, and precomputed FCU\footnote{Flight Control Unit} outputs are not used for correction. Consequently, the discrepancy represents prediction error relative to the sensor-based estimate, rather than absolute position error.

The observation must occur after the upper bound of the transmission interval and no later than two seconds after the source timestamp. \texttt{MANUAL\_CONTROL/TIMESYNC}, message ordering, and Receipt provide the basis for alignment. Prediction is interpolated to the observation time, and the origin is not shifted to conceal latency. A consistent state in NED is accepted only when ARM, mode, Intake sample, four-component command, and transmission evidence match. The residual is defined as:

\[
\mathbf r_k
=
\begin{bmatrix}
\mathbf p_k^{MESA}\\
\mathbf v_k^{MESA}
\end{bmatrix}
-
\hat{\mathbf x}
\left(
t_k
\mid
t_{\mathrm{source}},
\mathbf u_k
\right).
\]

In this equation, \(\mathbf r_k\) represents position discrepancy (meters) and velocity discrepancy (meters per second), \(\mathbf p_k^{MESA}\) and \(\mathbf v_k^{MESA}\) are the estimates, and \(\hat{\mathbf x}\) is the time-aligned prediction under command \(\mathbf u_k\), originating at \(t_{\mathrm{source}}\). Furthermore, \(t_k\) is the observation time (seconds). Only the velocity component is used for bounded disturbance correction:

\[
\mathbf b_{k+1}
=
\operatorname{clip}\!\left(
\mathbf b_k
+
0.06
\operatorname{clip}\!\left(
\frac{\mathbf r_{v,k}}
{\max(0.1,t_k-t_{\mathrm{source}})},
-2,2
\right),
-2,2
\right).
\]

In this equation, \(\mathbf b\) denotes the model acceleration disturbance (meters per second squared), \(\mathbf r_v\) denotes velocity discrepancy (meters per second), and \texttt{clip} denotes component-wise limiting within bounds of \(\pm2\,\mathrm{m/s^2}\). Time quantities are expressed in seconds. This correction neither distinguishes the source of error among estimation, wind, dynamics, and latency nor guarantees convergence. The position residual is not fed back into MESA-NAV. Independent MicroMet correction, navigation covariance adaptation, and comprehensive uncertainty propagation have not been implemented.

\subsection{Software Configuration and Reproducibility}

The Native test relies on the \texttt{TDDT\_phoenixdrone\_navigation\_GPSdenied} bundle under Linux, Python 3.13, MAVLink 2, ArduPilot SITL, and JSBSim \cite{ref26_nn44nniri_tddt_phoenix}. The package includes Python modules, C++ source code, three FMUs\footnote{FMU (Functional Mock-up Unit): A standardized, portable software unit that packages a simulation model and its interfaces for data exchange and co-simulation execution across different systems.}, the Phoenix XML model, ArduPilot parameters, TDOC configurations, DEM, atmospheric data, CRP/SARG/TEP knowledge and KStore, scripts, and evidence. The evaluated Test 20 software configuration, including its source identifier, execution profile, session record, and SHA-256 provenance manifests, is documented in the accompanying reproducibility repository \cite{ref26_nn44nniri_tddt_phoenix}.

The \texttt{environment\_fixed.sh} script configures paths, while \texttt{requirements.txt} specifies dependencies for TDOC and the layers\cite{ref26_nn44nniri_tddt_phoenix}. FlightStack and GUI are installed separately. Reproducibility requires compatible versions and datasets, rebuilding FMUs and modified models, installing JSBSim/ArduPilot XML files and parameters, and preparing atmospheric data and maps. Once the prerequisites are satisfied, TDOC activates the simulator, environment interface, GUI, and MAVProxy. Offline knowledge is recovered from three compatible training cycles or Handoff, but it does not authorize flight commands. The simulator clock, reception time, atmospheric Epoch, and TCP/UDP paths must be configured separately and consistently.

\section{Experimental Environment and Evaluation Methodology}

This section defines the experimental environment, scenarios, evaluation metrics, validation procedures, and comparison methods.

\subsection{Direct Flight Simulation Environment and Geographic Test Area}

The experiments were conducted near Boulder, Colorado, USA, with an origin at (\(40.010^\circ\) N, \(105.30^\circ\) W) and a destination at (\(40.025^\circ\) N, \(105.28^\circ\) W). The geodesic distance between these points was \(2385.231\,\mathrm{m}\). The TIFF raster, represented in UTM\footnote{Universal Transverse Mercator} Zone 13 North (EPSG:32613), comprised a \(343\times342\) grid with a spatial resolution of approximately \(10\,\mathrm{m}\) and an area of \(11.68\,\mathrm{km^2}\). The raster layers included elevation, slope, aspect, roads, buildings, and natural and artificial obstacles. Raster elevations at the two points were approximately \(1800\,\mathrm{m}\) and \(1636\,\mathrm{m}\), respectively. However, the vertical datum was not verified, and grid resolution does not constitute a measure of vertical accuracy.

ArduPilot managed flight control and actuator outputs, while JSBSim simulated the Phoenix dynamics independently of WingMPET. EKF3\footnote{Extended Kalman Filter 3} and QHOVER were enabled, whereas real/simulated GNSS and absolute horizontal position aiding remained disabled. Using versioned meteorological data and the DEM, the Environment Adapter verified the application of horizontal wind through parameter readback. However, the application of vertical wind, temperature, and pressure, as well as the correspondence between all MicroMet outputs and the actual disturbances, was not verified.

The GUI served as an interface for either the Digital Pilot or a gamepad; simultaneous use was not permitted. Before autonomous operation, manual control using a Logitech F310 was tested in JSBSim. Six windows associated with ArduPilot, the JSBSim observer, Environment Adapter, GUI (Fig. \ref{fig:phoniex_gui}), and MAVProxy Map/Console (Fig. \ref{fig:phoniex_map}, Fig. \ref{fig:phoniex_console}) displayed the trajectory, ARM state, flight mode, and telemetry. Data freshness was monitored over local TCP/UDP connections and MAVLink 2. The target TDOC execution rate was \(10\,\mathrm{Hz}\), the command service rate was \(30\,\mathrm{Hz}\), and MicroMet was invoked approximately every \(0.5\,\mathrm{s}\). The requested rates for IMU, airspeed/servo, landing state, and SIM\_STATE were \(100\), \(20\), \(5\), and \(10\,\mathrm{Hz}\), respectively. These rates do not represent the JSBSim simulation or display update rates.

\subsection{Flight Mission Scenarios}

The tests included local hovering and landing, as well as an origin-to-destination mission. The complete sequence comprised \texttt{vertical\_takeoff}, \texttt{cruise\_vtol}, \texttt{vertical\_landing}, \texttt{landing\_flare}, \texttt{contact\_settle}, and \texttt{DISARM}. However, an independent hover was performed only in the local test. The initial conditions consisted of the Phoenix positioned vertically on the ground with its motors inactive, a known origin and heading, QHOVER mode, a \(60\,\mathrm{s}\) stationary calibration, and versioned meteorological conditions. Real/simulated GNSS, vision, and odometry were disabled for navigation. JSBSim ground truth was used exclusively as an evaluation reference, and normal ARM operation was confirmed through telemetry.

In the local test, the destination coincided with the origin, the target altitude was \(3\,\mathrm{m}\), and the requested hover duration was \(3\,\mathrm{s}\). Entry into hover required an estimated altitude greater than the target altitude minus \(0.5\,\mathrm{m}\) and an absolute vertical speed below \(0.5\,\mathrm{m/s}\). The hover constraints consisted of an estimated horizontal distance of at most \(5\,\mathrm{m}\), an altitude error of at most \(1\,\mathrm{m}\), and a three-dimensional speed below \(1\,\mathrm{m/s}\). In the implementation, verification of these conditions continued for \(3\,\mathrm{s}\) beyond the requested duration; any constraint violation reset the counter.

The reference altitude and speed for the trajectory were \(10\,\mathrm{m}\) above the origin and \(6\,\mathrm{m/s}\), respectively. The \texttt{cruise\_vtol} phase required an estimated altitude above \(9.5\,\mathrm{m}\) and an absolute vertical speed below \(0.5\,\mathrm{m/s}\); no transition to fixed-wing flight was performed. Descent was initiated when the estimated horizontal distance was below \(3\,\mathrm{m}\) and the horizontal speed was below \(0.8\,\mathrm{m/s}\). Estimated proximity did not constitute physical acceptance of the destination. The NED descent reference was \(0.8\,\mathrm{m}\) below the origin. Below an estimated altitude of \(1.5\,\mathrm{m}\), \texttt{landing\_flare} set the horizontal and rotational commands to zero and limited the descent rate to \(0.25\,\mathrm{m/s}\). The mission-control logic assumed ground approximately level with the origin. The DEM was used for geographic visualization rather than terrain-constrained control, and the plotted flight path was not adjusted to follow the DEM surface; consequently, terrain-referenced landing clearance and equivalence between the simulated contact surface and DEM elevations were not independently established.

\begin{figure}[htbp]
\centering
\includegraphics[width=0.99\linewidth,alt={Mission-state flowchart showing initialization, calibration, arming, vertical takeoff, alternative hover and full-mission branches, descent, landing flare, contact verification, ground stabilization, and disarming. Failure conditions lead to test termination, while successful completion enables independent simulator validation.}]{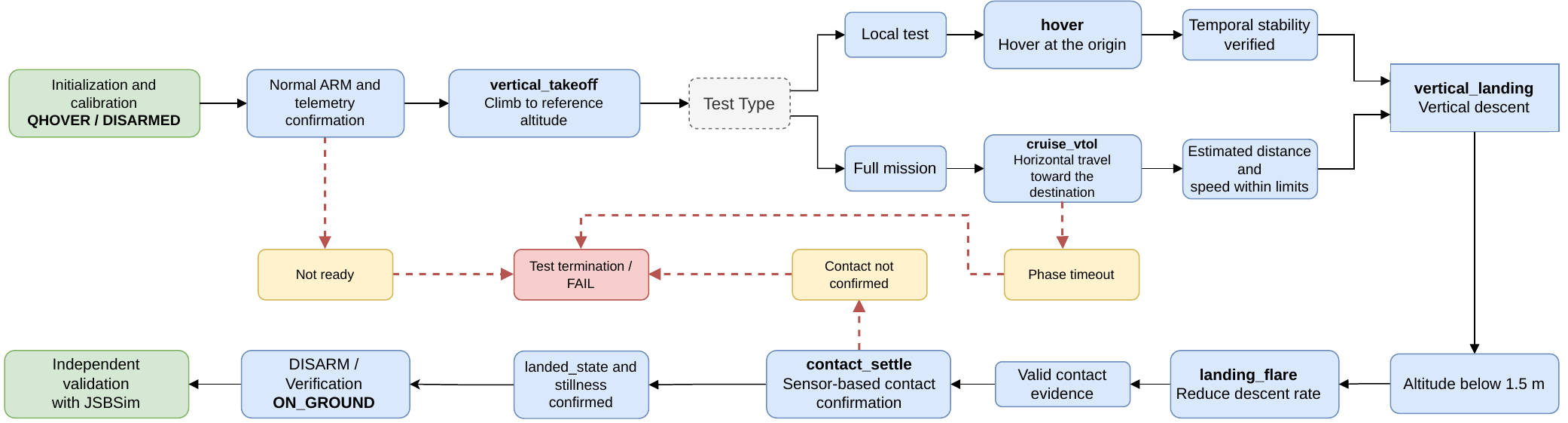}
\caption{Mission-phase execution and termination workflow for the Phoenix VTOL in Native simulation. Following initialization, calibration, and confirmed arming, vertical takeoff leads to either local hovering or the complete destination-oriented mission. Both branches converge on vertical descent, landing flare, sensor-based contact confirmation, ground-state verification, and DISARM. Readiness failure, phase expiration, or unconfirmed contact leads to test termination. Independent JSBSim-based validation follows the successful completion of the landing sequence.}
\label{fig:mission_phase_execution}
\end{figure}

The \texttt{contact\_settle} phase required valid barometric evidence of stillness or an inertial impact near the ground. Subsequently, QSTABILIZE mode and a zero command were applied. Landing completion required \texttt{landed\_state=1}, an estimated altitude below \(0.8\,\mathrm{m}\), a vertical speed below \(0.4\,\mathrm{m/s}\) for at least \(2\,\mathrm{s}\), and confirmed DISARM. The independent physical destination-error threshold was \(25\,\mathrm{m}\), the time limit for each phase was \(900\,\mathrm{s}\), and the mission limit was \(20000\) cycles. The operational sequence of the local hovering test and the complete VTOL mission, including phase-transition conditions, landing verification, failure branches, and independent post-mission validation, is illustrated in Figure \ref{fig:mission_phase_execution}.

Three attempts were recorded: Test 18, which ended in an unsuccessful landing; local Test 19, which achieved valid hovering and landing; and Mission 20, which underwent independent validation. The test identifier, configuration and meteorological versions, mission phases, commands, telemetry, and termination were recorded. Independent no-wind, headwind, tailwind, or crosswind tests, as well as repetitions under identical conditions, were not verified. The three route-evaluation sessions ran on 4, 5, and 6 November 2026; the first was Mission 20 (attempt 20), distinct from preliminary Test 18 and local Test 19. The evaluation was based on versioned meteorological conditions and the wind actually applied. Because the sessions were not matched-condition repetitions, their accuracy differences cannot be attributed to a specific factor.

\subsection{Performance Evaluation Metrics}

The evaluation metrics are based on state estimates, JSBSim, and TDOC evidence. Alignment of time, origin, NED coordinates, and altitude reference, together with the recording of units, provenance, test identity, and valid samples, is mandatory. SIM\_STATE is not incorporated into state estimation.

Horizontal position error and RMSE\footnote{Root Mean Square Error} are calculated as follows:

\[
e_{p,k}=\sqrt{(\hat p_{N,k}-p^{JSB}_{N,k})^2+(\hat p_{E,k}-p^{JSB}_{E,k})^2},
\quad RMSE_p=\sqrt{\frac{1}{|\mathcal V|}\sum_{k\in\mathcal V}e_{p,k}^{2}}.
\]

In these expressions, \(\hat p_N,\hat p_E\) denote the estimated coordinates, and \(p_N^{JSB},p_E^{JSB}\) represent the reference coordinates in meters. The index \(k\) identifies the sample, and \(\mathcal V\) denotes the set of valid, time-aligned samples. Both metrics are expressed in meters. The median, 95th and 99th percentiles, maximum, and RMSE are reported when raw data are available. Velocity error is defined as the difference between ground-velocity vectors, expressed in meters per second. These metrics have no independent acceptance thresholds, and the seventh-layer residual is not a substitute for spatial position error.

Destination error is defined as the horizontal WGS84\footnote{World Geodetic System 1984} geodesic distance between the destination and the last valid physical JSBSim position after landing. The acceptance threshold is \(25\,\mathrm{m}\), whereas an estimated distance below \(3\,\mathrm{m}\) only triggers descent. Vertical error requires a verified vertical datum.

Processing time is the difference between the end and start of a TDOC cycle, measured using the host's monotonic clock. Decision latency is the difference between the decision time and the input validity time. Both are expressed in milliseconds. The median, 95th and 99th percentiles, maximum, and proportion of cycles exceeding \(100\,\mathrm{ms}\) are reported. Decision, transmission, Receipt, and observation latencies are treated separately. Their comparison requires a shared clock or valid temporal alignment. The fast Receipt deadline is \(120\,\mathrm{ms}\), and command validity is \(500\,\mathrm{ms}\), without a real-time guarantee.

Control continuity and exchange validity are defined, respectively, as the ratios of eligible cycles and transactions supported by valid evidence:

\[
R_{\mathrm{loop}}=\frac{\sum_{k\in\mathcal E_L}I_k^L}{|\mathcal E_L|},
\qquad R_{\mathrm{tx}}=\frac{\sum_{j\in\mathcal E_T}I_j^T}{|\mathcal E_T|}.
\]

In these expressions, \(R_{\mathrm{loop}},R_{\mathrm{tx}}\) are dimensionless ratios; \(\mathcal E_L,\mathcal E_T\) denote the eligible sets; \(k,j\) are the cycle and transaction indices; and \(I^L,I^T\) are binary validity indicators. The former metric evaluates the traceable chain of state, prediction, decision, command, and feedback, including feedback received in a subsequent cycle. The latter evaluates the transaction identifier, timestamp, Receipt, and transmission evidence. Software-level acceptance does not constitute transmission or a physical response.

Mission success requires completion of the mission phases, a physical destination error of no more than \(25\,\mathrm{m}\), valid contact, and confirmed \texttt{ON\_GROUND} and \texttt{DISARM} states. A final speed below \(0.4\,\mathrm{m/s}\), a center-of-mass height difference below \(0.5\,\mathrm{m}\), and an attitude deviation from the vertical below \(25^\circ\) are also required. GNSS status and evidence validity are checked. Success, failure, and early termination are recorded separately; non-identical tests do not constitute statistical repetitions.

\subsection{Independent Validation Method}

Validation is based on separating JSBSim ground truth from operational processes. Ground-truth geographic position, altitude, three-axis velocity, and quaternion are recorded exclusively for visualization and post hoc evaluation. Informational independence does not imply physical independence of the reference from the sensor-data generator.

Only valid, provenance-tracked telemetry from inertial, magnetic, pressure, and airspeed sensors is provided to MESA-NAV. JSBSim ground-truth position and velocity, the controller's ready-made position, and its estimated orientation are excluded. Real/simulated GNSS, horizontal position aiding, vision, and odometry are disabled, and GNSS settings are verified through readback. ArduPilot may use its internal state for control, but it does not provide that state to MESA-NAV as a positional observation.

The origin, destination, DEM, and offline knowledge constitute prior information; they do not reset the estimated position. The true position displayed by the visualization interface is not fed back into the shared state or decision-making process. The Environment Adapter may use ground-truth position to apply wind to JSBSim, but it does not generate synthetic anemometer observations. The seventh layer performs only limited WingMPET adaptation using sensor-based observations obtained after command execution.

The WGS84 reference is converted into local north--east coordinates and aligned with the estimates according to source time, provenance, and coordinate system. A future sample does not replace a past observation, missing data are not assumed to be zero, and an invalid vertical datum is not accepted. Errors are calculated according to Section~5.3. The audit examines data provenance, ground-truth exclusion flags, the shared-state hash, transaction identifiers, transmission evidence, post-transmission observations, and single-use feedback consumption. Native reports demonstrate that contractual checks were passed. However, no intervention test was conducted in which ground truth was deliberately removed or modified and its lack of influence on decisions was verified. Therefore, the absolute absence of information leakage and the validity of real-world flight have not been demonstrated.

\subsection{Comparison Method with Reference Approaches and Component Ablation Tests}

TDDT--Phoenix serves as the comparison baseline, but no matched-condition comparison against a reference approach has been performed. The only offline, three-sample WingMPET test compared MicroMet wind against zero wind under identical air-command inputs. This test evaluates model sensitivity and does not provide evidence of improved mission performance.

In future comparisons, baseline inertial navigation will be evaluated against MESA-NAV, reactive guidance against predictive guidance, and guidance based on a simplified model against the complete system. The origin, mission, aircraft, sensors, meteorological conditions, prior information, disabled GNSS, constraints, and computational budget must remain identical, and any differences must be reported. The metrics and acceptance conditions defined in Section~5.3 and the independent reference specified in Section~5.4 must also remain unchanged.

In ablation tests, MESA-NAV will be replaced by a baseline estimator, WingMPET by a simplified model, and MicroMet by fixed meteorological conditions, without changing the physical wind applied in JSBSim. Restricting coupling differs from removing a digital twin. When feedback adaptation is disabled, observation and recording must continue. SARG ablation is also proposed. Removing a component without a compatible replacement merely evaluates execution failure.

Comparison requires paired executions in which only one factor is changed, an independent reference, and the recording of success, failure, or non-comparability. Differences in evaluation metrics indicate the effect of the intervention under the same conditions. However, attributing an independent contribution requires repetitions under identical conditions and control of confounding factors. The existing tests are not paired; therefore, neither the experimental superiority of the architecture nor the quantitative contributions of individual digital twins have been demonstrated.

\section{Experimental Results}

This section presents evidence of mission execution, navigation accuracy, processing time, integration, and safety from software-in-the-loop simulation tests.

\subsection{Results of Complete Mission Execution}

In the documented development tests of TDDT–Phoenix, with a known origin and online GNSS disabled, Test 18 failed because contact was not detected and landing was not completed, whereas local Test 19 successfully verified hovering and landing. The subsequent full-route Test 20, conducted on 4 November 2026, was accepted (Sections 5.1 and 5.2). These tests had different objectives and were not repetitions of an identical protocol. Separately, 145 selected software tests passed; these were not additional flight missions. The geodesic reference distance from origin to destination was \(2385.23\,\mathrm{m}\). Test 20 recorded 4441 control cycles over \(443.35\,\mathrm{s}\), and the physical destination error was \(21.61\,\mathrm{m}\), below the \(25\,\mathrm{m}\) threshold. Takeoff, contact, ground stability, DISARM, the transaction chain, and the unique consumption of 1646 feedback items were also confirmed.

\begin{figure}[htbp]
\centering
\includegraphics[width=0.95\linewidth,alt={Two-panel mission-results chart comparing physical and estimated distances to the destination and summarizing horizontal navigation errors. The third test ends 31.62 m from the destination, exceeding the 25 m acceptance limit, while the estimated distance is 6.42 m. Error statistics include a median of 8.33 m and a maximum of 29.71 m.}]{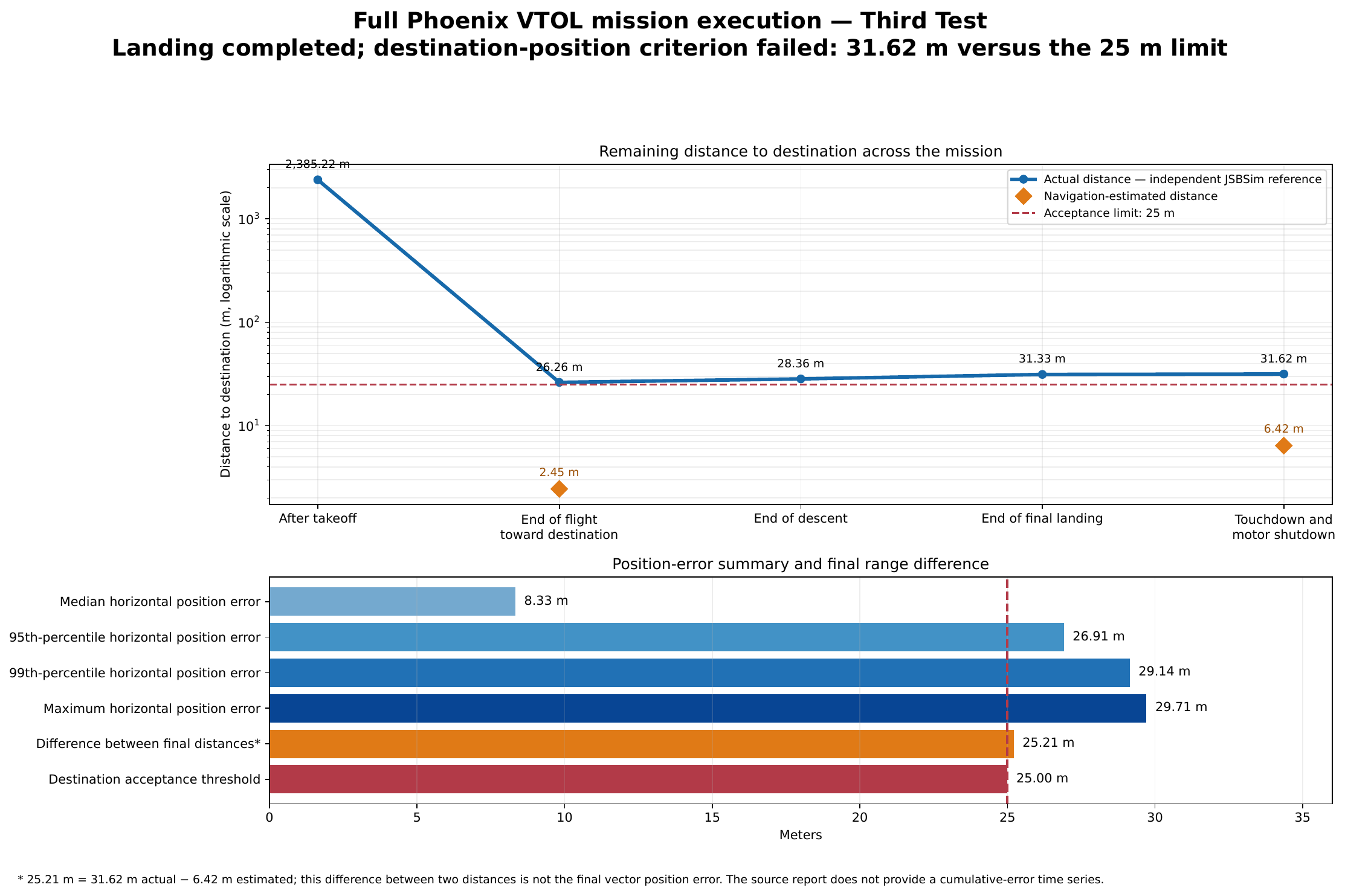}
\caption{Mission outcome and horizontal navigation-error summary for the third test. The upper panel compares physical JSBSim and MESA-NAV-estimated distances to the destination at selected mission checkpoints, using a logarithmic scale and the 25 m acceptance threshold. The lower panel summarizes horizontal position-error statistics and the difference between final destination distances. This scalar distance difference is distinct from the terminal vector position-estimation error.}
\label{fig:mission_outcome_horizontal_navigation-error_summary}
\end{figure}

In the third independent test, 4455 cycles were recorded over \(444.78\,\mathrm{s}\). The durations of takeoff, VTOL travel, descent, final landing, and contact stabilization were \(9.99\), \(407.43\), \(14.27\), \(7.90\), and \(5.19\,\mathrm{s}\), respectively, and the physical takeoff altitude reached \(9.67\,\mathrm{m}\). The actual distances to the destination at the end of travel, the end of descent, and after contact were \(26.26\), \(28.36\), and \(31.62\,\mathrm{m}\), respectively. Fixed-wing flight did not occur. Figure \ref{fig:mission_outcome_horizontal_navigation-error_summary} summarizes the third test's physical and estimated destination distances across mission checkpoints and the associated horizontal position-error statistics.

Sensor data and the physical response from Layer 1 supplied the three twins in Layer 2, while Layer 3 constructed 4455 atomic shared states. Prediction, decision-making, and command acceptance were recorded in every cycle. Layer 7 accepted 1527 feedback items, of which 1526 unique items were consumed in subsequent cycles. Despite confirmed contact, ground state, and motor shutdown, the final error of \(31.62\,\mathrm{m}\) exceeded the allowable limit; therefore, controlled termination did not constitute mission success. The evidence supports the operational loop and the separation of JSBSim ground truth from navigation, but does not demonstrate the realization of a complete dynamic network, guaranteed global feasibility, hardware flight, or superiority over reference methods.

\subsection{Navigation and Position Estimation Accuracy}

\begin{figure}[htbp]
\centering
\includegraphics[width=0.95\linewidth,alt={Two horizontal trajectory plots comparing the physical JSBSim flight path and MESA-NAV estimated path in local north–east coordinates. The destination close-up highlights endpoint divergence and the 25 m acceptance radius, with final physical and estimated distances of 31.62 m and 6.42 m.}]{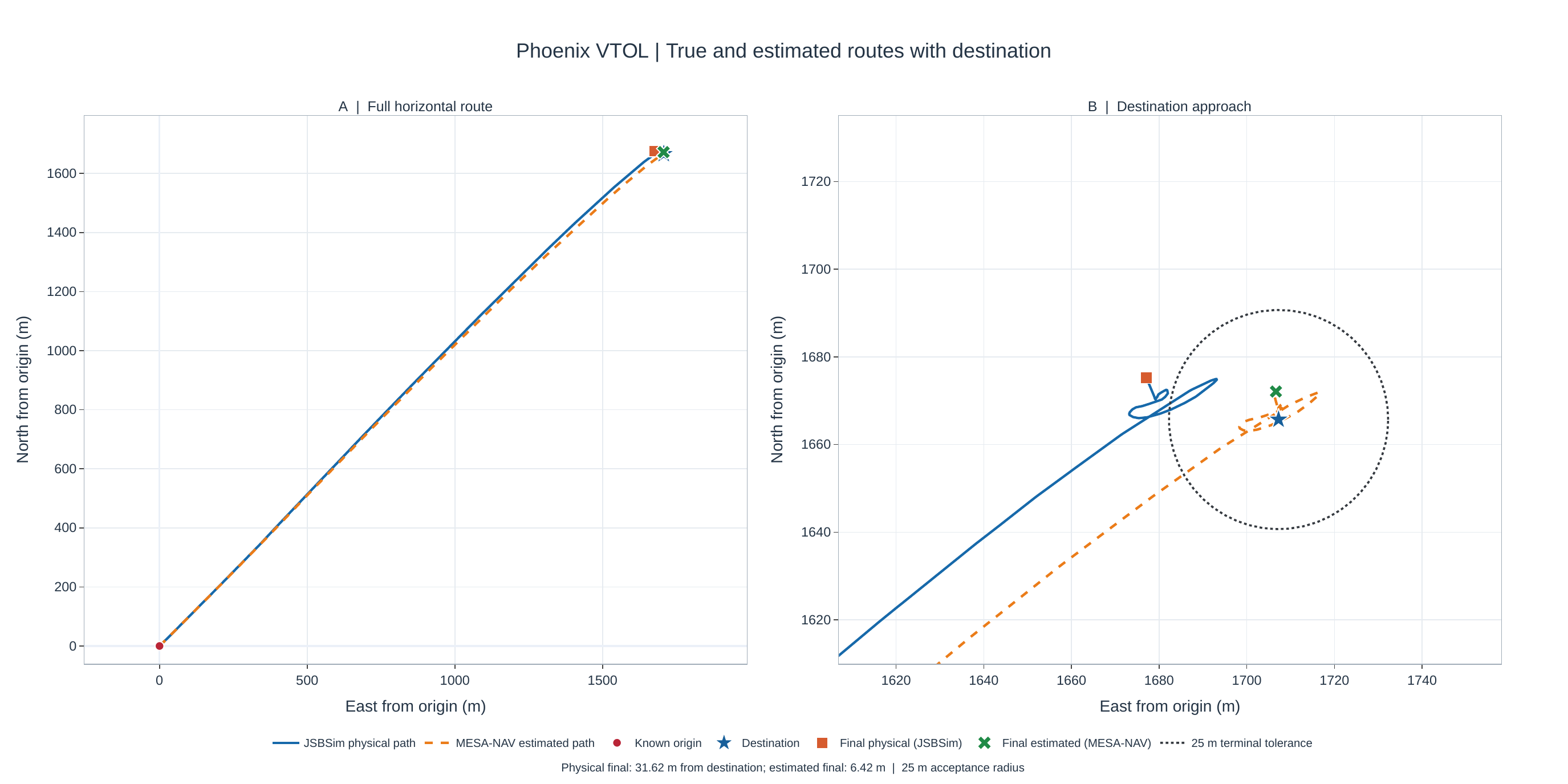}
\caption{Comparison of the physical and estimated horizontal flight paths of Phoenix VTOL during the third test. Panel A presents the trajectories in local east–north coordinates, while Panel B magnifies the destination approach and displays the 25 m acceptance radius. JSBSim provides the independent physical reference, and MESA-NAV provides the navigation estimate. The final physical and estimated distances to the destination are 31.62 m and 6.42 m, respectively. No post-processing alignment of the trajectories was applied.}
\label{fig:physical_estimated_horizontal_flight_paths}
\end{figure}

In the third test, the MESA-NAV position was compared with the independent JSBSim reference using 4455 time-aligned records over \(444.778\,\mathrm{s}\) (Sections 5.3 and 5.4). Online GNSS, vision, and odometry were disabled; reference independence was informational rather than physical. The median, mean, RMSE, 95th percentile, 99th percentile, and maximum horizontal position errors were \(8.33\), \(10.4148\), \(13.6649\), \(26.91\), \(29.14\), and \(29.71\,\mathrm{m}\), respectively. The terminal error was \(29.714\,\mathrm{m}\); vertical and three-dimensional errors were not evaluated because a common vertical datum had not been verified. Figure \ref{fig:physical_estimated_horizontal_flight_paths} compares the JSBSim reference trajectory with the MESA-NAV estimated trajectory, including their divergence near the destination.

\begin{figure}[htbp]
\centering
\includegraphics[width=0.90\linewidth,alt={Time-series plot of horizontal position-estimation error over 4,455 synchronized samples. Error increases during the mission and reaches approximately 29.71 m at the final sample. Vertical markers indicate mission phases; the mean, RMSE, and 95th-percentile errors are 10.41 m, 13.66 m, and 26.91 m.}]{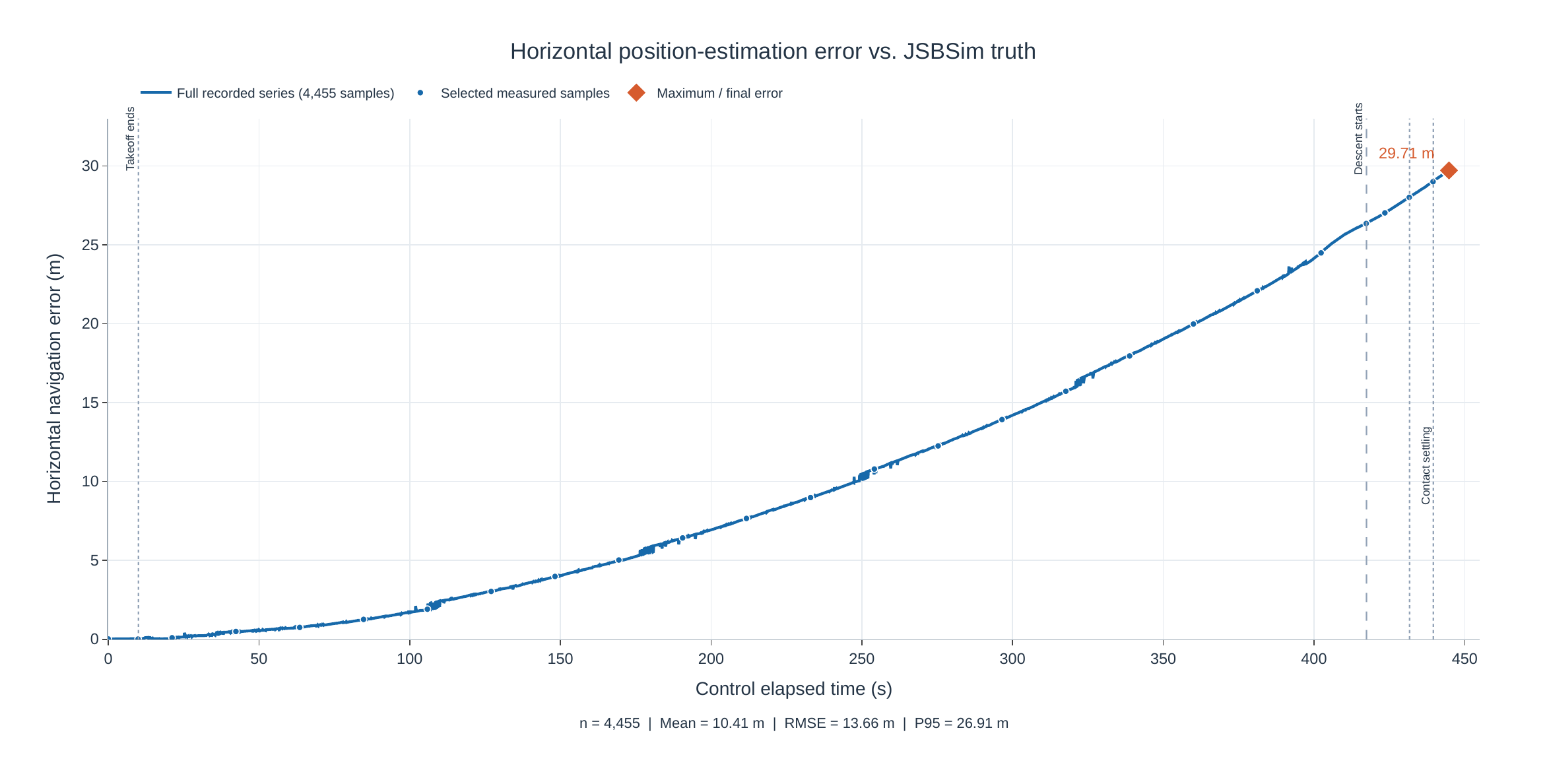}
\caption{Temporal evolution of horizontal position-estimation error relative to the JSBSim reference during the third test. The plot includes 4,455 time-aligned records, selected mission-phase markers, and the terminal maximum error. The mean error, root-mean-square error, and 95th-percentile error are 10.41 m, 13.66 m, and 26.91 m, respectively. The maximum error reaches approximately 29.71 m at the final recorded sample.}
\label{fig:temporal_evolution_horizontal_position-estimation_error}
\end{figure}

The mean errors during takeoff, travel, descent, final landing, and contact stabilization were \(0.013\), \(9.49\), \(27.15\), \(28.51\), and \(29.38\,\mathrm{m}\), respectively. The error reached \(26.35\,\mathrm{m}\) at the end of travel and did not decrease during descent. The first crossings of the \(10\) and \(20\,\mathrm{m}\) thresholds occurred at \(247.45\) and \(360.40\,\mathrm{s}\), respectively; the maximum error also occurred at the final sample. The time-integrated error was calculated using the trapezoidal relation:

\[
A_e \approx
\sum_{k=1}^{n-1}
\frac{e_{p,k}+e_{p,k+1}}{2}
\,\Delta t_k.
\]

In this relation, \(A_e\) denotes accumulated error (meter-seconds), \(e_{p,k}\) is the horizontal error of sample \(k\) (meters), \(\Delta t_k\) is the time interval between two samples (seconds), and \(n\) is the number of valid samples. Its value was \(4631.87\,\mathrm{m\,s}\); the net terminal drift was \(29.72\,\mathrm{m}\), which is not equivalent to either RMSE or time-integrated error. The accumulation of horizontal position-estimation error throughout the mission, together with the corresponding flight-phase transitions, is shown in Figure \ref{fig:temporal_evolution_horizontal_position-estimation_error}.

\begin{figure}[htbp]
\centering
\includegraphics[width=0.99\linewidth,alt={Two-panel comparison of navigation accuracy across mission phases and physical versus estimated destination distances. Horizontal estimation errors increase toward landing. Although the vehicle temporarily approaches within 25 m of the destination, its final physical distance is 31.62 m, and the destination criterion fails.}]{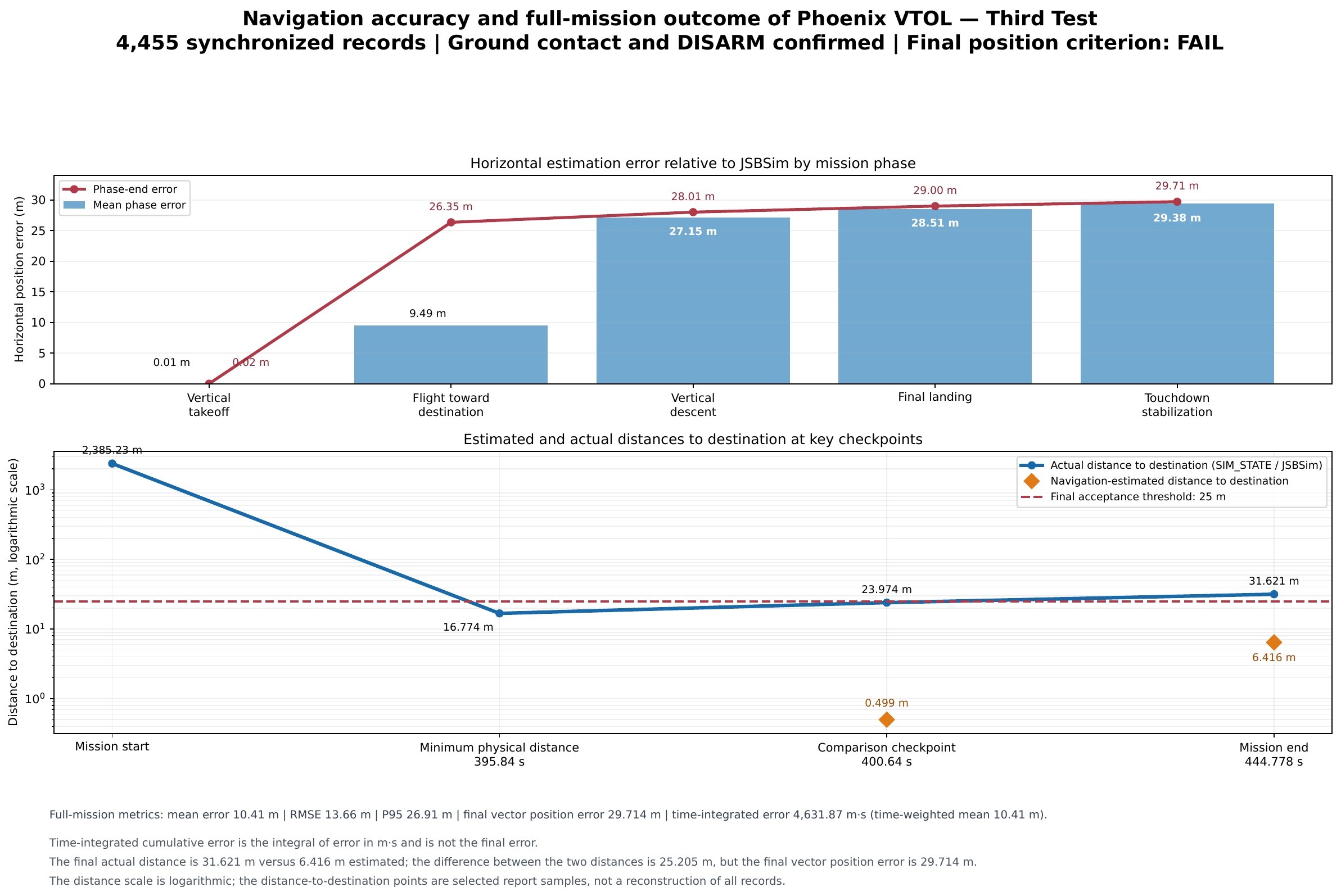}
\caption{Phase-dependent navigation accuracy and destination-position outcome of the Phoenix VTOL in the third test. The upper panel compares mean and phase-end horizontal position errors, while the lower panel contrasts physical JSBSim and MESA-NAV-estimated distances to the destination at selected checkpoints. Although touchdown and motor shutdown were confirmed, the final physical distance exceeded the 25 m acceptance threshold. The difference between final destination distances must not be interpreted as the terminal vector position error.}
\label{fig:navigation_accuracy_destination-position_outcome}
\end{figure}

The minimum actual distance to the destination was \(16.77\,\mathrm{m}\) at approximately \(395.8\,\mathrm{s}\). At \(400.64\,\mathrm{s}\), the estimated distance was \(0.50\,\mathrm{m}\), whereas the actual distance was \(23.97\,\mathrm{m}\). After contact, these distances reached \(6.42\) and \(31.62\,\mathrm{m}\), respectively; the simultaneous vector position discrepancy was \(29.71\,\mathrm{m}\), which is not equivalent to the difference between the two destination distances. Therefore, temporary entry into the \(25\,\mathrm{m}\) radius did not guarantee a valid landing. The first accepted route test, with a destination error of \(21.61\,\mathrm{m}\), was not a repetition of the third test under identical conditions. Offline data, comparison with reference navigation, and twin-ablation tests did not demonstrate a reduction in closed-loop error. The data indicate horizontal drift, not statistical superiority of the method. Figure \ref{fig:navigation_accuracy_destination-position_outcome} summarizes the phase-dependent growth of navigation error and the divergence between estimated and actual destination distances at critical mission checkpoints.

\subsection{Real-Time Computational Performance}

The decision cycle, with a target frequency of \(10\,\mathrm{Hz}\), had a nominal period of \(100\,\mathrm{ms}\); the \(120\,\mathrm{ms}\) acknowledgment deadline and \(500\,\mathrm{ms}\) command validity were separate criteria. In the third test (4455 cycles, \(444.78\,\mathrm{s}\)), the mean, median, 95th percentile, 99th percentile, and maximum processing times were \(40.60\), \(36.96\), \(86.27\), \(106.88\), and \(181.71\,\mathrm{ms}\), respectively. In total, 74 cycles (\(1.66\%\)), 13 cycles, and one cycle exceeded the \(100\), \(120\), and \(150\,\mathrm{ms}\) thresholds, respectively.

\begin{figure}[htbp]
\centering
\includegraphics[width=0.99\linewidth,alt={Four-panel computational performance chart showing cycle-time distribution, phase-dependent processing times, comparisons across three simulation sessions, and data-source versus command-receipt timing. The third test records a mean of 40.60 ms, P95 of 86.27 ms, P99 of 106.88 ms, and maximum of 181.71 ms against a nominal 100 ms cycle period.}]{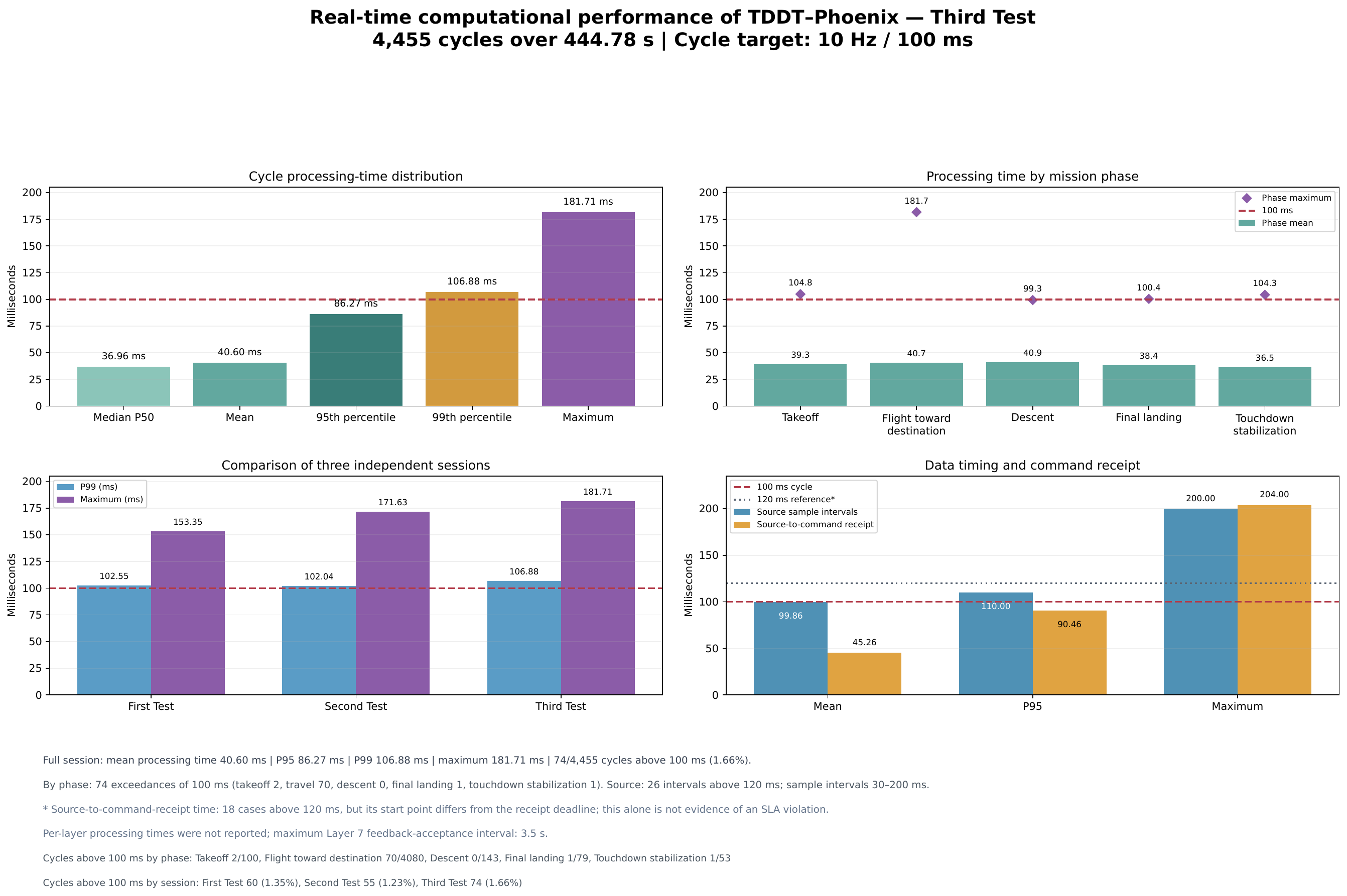}
\caption{Computational timing performance of TDDT–Phoenix in Native simulation. The panels present processing-time statistics for the third test, mean and maximum processing times by mission phase, comparisons across three recorded sessions, and source-sample timing versus source-to-command-receipt intervals. The nominal 100 ms cycle period is shown for comparison. The receipt-related timing measures have different reference points and do not independently demonstrate acknowledgment-deadline violations or hard real-time guarantees.}
\label{fig:computational_timing_performance}
\label{fig:Three-dimensional_visualization_Phoenix_VTOL_mission_path}
\caption*{\footnotesize\textit{Note:} 50th Percentile (P50): The value at or below which 50\% of observations fall, corresponding to the median. 95th Percentile (P95): The value at or below which 95\% of observations fall. 99th Percentile (P99): The value at or below which 99\% of observations fall.}
\end{figure}

Across 4454 source-time intervals, the mean, 95th percentile, and maximum were \(99.86\), \(110\), and \(200\,\mathrm{ms}\), respectively. The source-to-command-acknowledgment interval had a mean of \(45.26\,\mathrm{ms}\), a 95th percentile of \(90.46\,\mathrm{ms}\), and a maximum of \(204\,\mathrm{ms}\). Exceeding \(120\,\mathrm{ms}\) in this interval does not necessarily indicate a violation of the acknowledgment deadline because the timing reference points differ. Figure \ref{fig:computational_timing_performance} summarizes the distribution of cycle processing times, phase-dependent computational delays, differences among recorded sessions, and source-to-command timing measurements.

In the other two tests, 4441 and 4458 cycles were recorded, with 95th-percentile processing times of \(81.83\) and \(82.05\,\mathrm{ms}\), respectively; 60 and 55 cycles exceeded \(100\,\mathrm{ms}\). Both missions were accepted, and a causal relationship between latency and rejection of the third test was not established. Execution of Layers 2 through 6 was recorded, but individual layer execution times, complete sensor-to-physical-actuator-response latency, worst-case execution time, processor load, and edge-processor performance were not measured. Therefore, the results indicate near-real-time performance with intermittent delays, rather than hard real-time operation; the offline report also does not establish readiness for online control.

\begin{figure}[htbp]
\centering
\includegraphics[width=0.99\linewidth,alt={Three-dimensional terrain visualization showing the Phoenix VTOL simulated mission trajectory over a colored digital elevation model. The plot includes local horizontal coordinates, terrain elevations referenced to MSL, and markers identifying the mission origin, destination, and selected flight events.}]{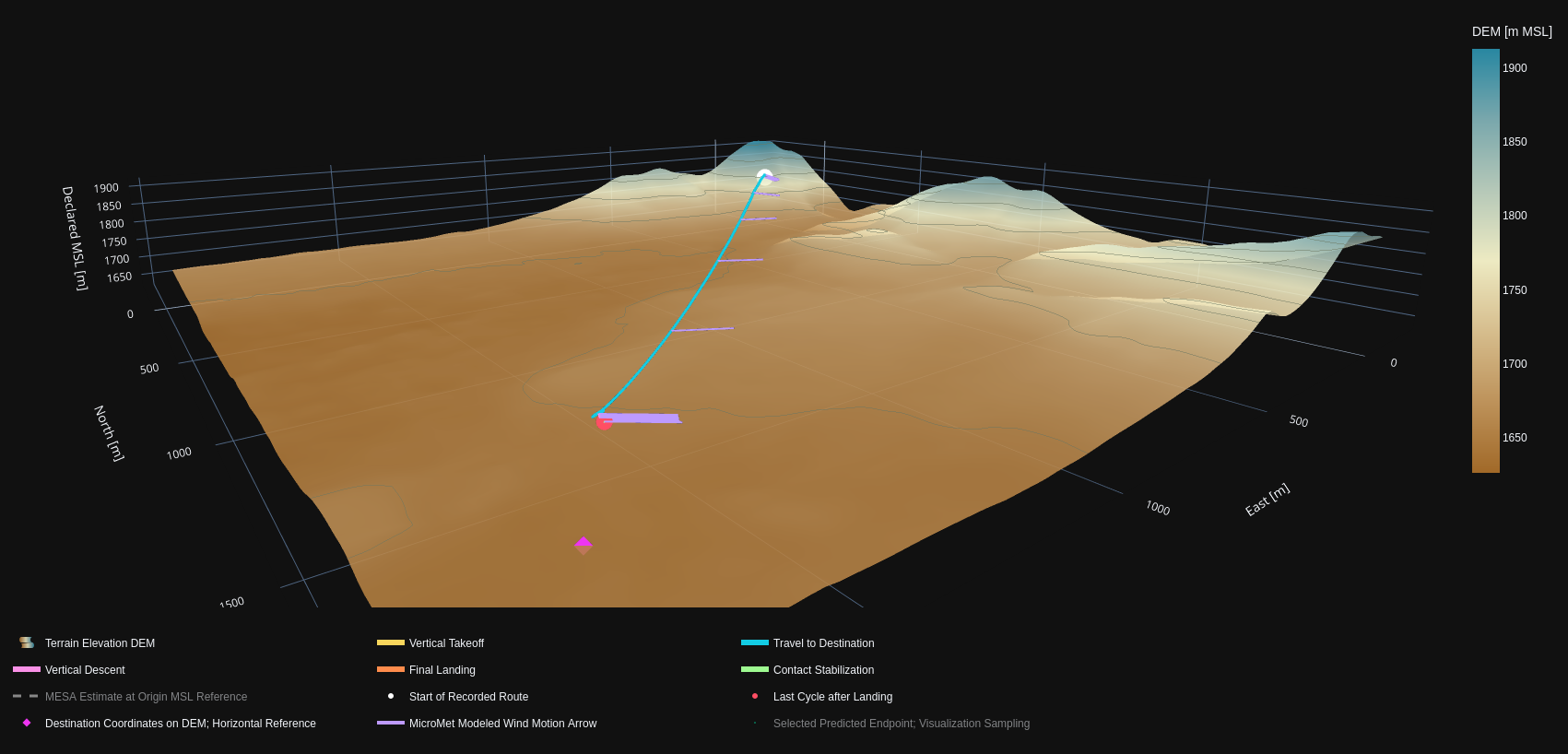}
\caption{Three-dimensional visualization of the Phoenix VTOL mission path over the regional digital elevation model. The DEM provides geographic context using local north–east coordinates and elevations with a declared MSL reference; equivalence of the vertical datum to the flight-state and physical-contact references was not independently certified. The physical trajectory was plotted from recorded simulator states without terrain draping or altitude adjustment to the DEM surface. Accordingly, this visualization does not establish terrain clearance, destination touchdown elevation, or terrain-related safety compliance.}
\label{fig:Three-dimensional_visualization_Phoenix_VTOL_mission_path}
\caption*{\footnotesize\textit{Note:}Mean Sea Level (MSL): A vertical reference datum used to express elevations relative to the long-term average sea surface level.}
\end{figure}

\subsection{Cross-Domain Integration and Feedback Performance}

During Native execution, WingMPET used the position and attitude estimated by MESA-NAV, wind and density provided by MicroMet, and actuator data for prediction; the initial meteorological wind served only as a navigation prior. Layer 3 aligned the information for prediction, decision-making, and execution. Source timestamps and identifiers, freshness, transaction identity, and packet hashes were checked. The maximum permitted age of meteorological data was three seconds, and their asynchronous generation did not require forced synchronization. Figure \ref{fig:Three-dimensional_visualization_Phoenix_VTOL_mission_path} provides a three-dimensional visualization of the simulated mission path over the regional DEM, illustrating the terrain context associated with the environmental model.

\begin{figure}[htbp]
\centering
\includegraphics[width=0.90\linewidth,alt={Cross-domain dataflow diagram and two statistical panels showing exchanges among the navigation, dynamics, and meteorological twins, shared-state coordination, execution, and feedback. The third test records 4,455 valid atomic packets, 22,275 fixed-topology relationship records, and 1,526 uniquely consumed feedback items out of 1,527 accepted.}]{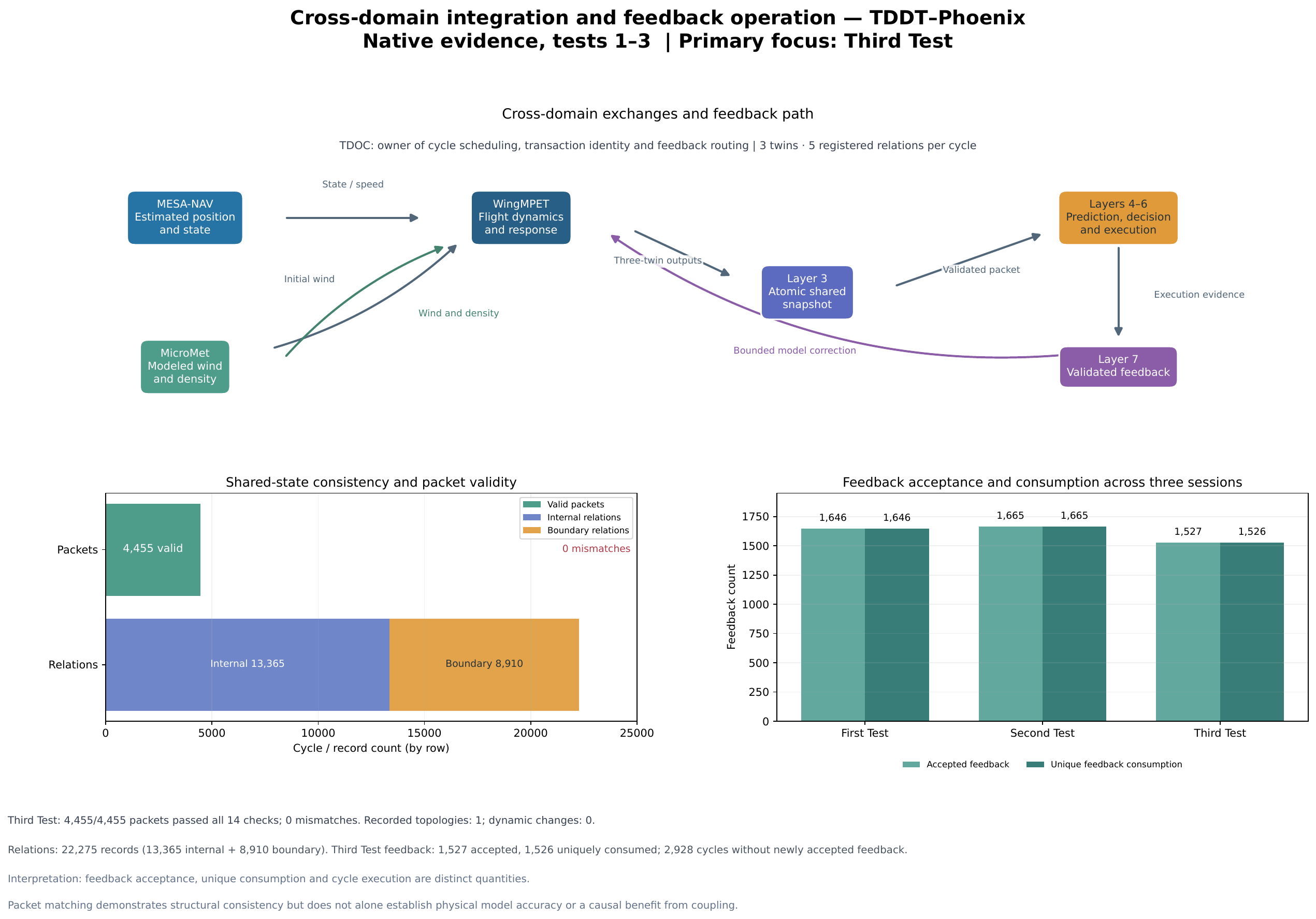}
\caption{Cross-domain integration and feedback operation in the Native TDDT–Phoenix implementation. The upper panel illustrates the directed exchanges among MESA-NAV, WingMPET, MicroMet, the shared-state layer, prediction and execution layers, and feedback processing. The lower panels summarize atomic-packet consistency, recorded relationships, and feedback acceptance and unique consumption across three sessions. The third test produced 4,455 valid atomic packets and 22,275 relationship records under a fixed topology. These observations establish contractual consistency and traceability, not physical model accuracy or the benefits of dynamic reconfiguration.}
\label{fig:cross-domain_integration_feedback_operation_Native_TDDT}
\end{figure}

In the third test, all 4455 atomic packets were consistent across 14 contractual checks covering hashes, the presence of the three twins, input correspondence, and decision identity; the 4458 packets from the second test were also consistent. In addition, 22275 directed connections were recorded under a fixed topology, comprising 13365 internal connections and 8910 boundary connections. This evidence confirms contractual consistency but does not establish the physical accuracy of the models or the computation of dynamic trust. Figure \ref{fig:cross-domain_integration_feedback_operation_Native_TDDT} summarizes the implemented cross-domain information exchanges, atomic shared-state consistency, and feedback acceptance and consumption across the recorded tests.

Feedback was accepted after verification of the test, transaction, command, hash, execution mode, and transmission/observation sequence. Subsequently, predictions were compared with sensor-based observations, and WingMPET disturbances were adjusted to a limited extent in the following cycle; JSBSim ground truth, MESA-NAV resetting, and independent MicroMet correction were not involved. In the second and third tests (5 and 6 November 2026), 1665 and 1527 feedback items were accepted, respectively, and 1665 and 1526 unique items were consumed; no consumption without a valid preceding record was observed. The absence of newly accepted feedback during 2928 cycles of the third test did not necessarily indicate a halt in decision-making. The adaptation center and the theoretical dynamic TDDT network \cite{ref09_amiri2026transdomain} were not implemented; furthermore, the causal contribution of each twin or the feedback mechanism cannot be established without ablation tests and comparisons under matched windy and windless conditions.

\subsection{Mission Reliability and Safety Evidence}

In the second and third tests, normal \texttt{ARM} was performed after calibration and \texttt{Pre-Arm} verification, and the five phases of takeoff, VTOL travel, descent, landing, and contact stabilization were completed. The first four phases were executed in \texttt{QHOVER}, and stabilization was performed in \texttt{QSTABILIZE}; fixed-wing transition was not tested. All 4458 and 4455 decisions were accepted at the Layer 6 input, but this acceptance does not independently establish the physical execution of every command, and a separate count of communication rejections was not reported.

\begin{figure}[htbp]
\centering
\includegraphics[width=0.99\linewidth,alt={Four-panel mission-reliability chart comparing flight-phase durations, destination accuracy, software-level command acceptance, and simulated energy estimates. The third test completes landing but exceeds the 25 m destination limit at 31.62 m. All 4,455 command inputs are accepted, while SITL and WingMPET energy integrals differ substantially.}]{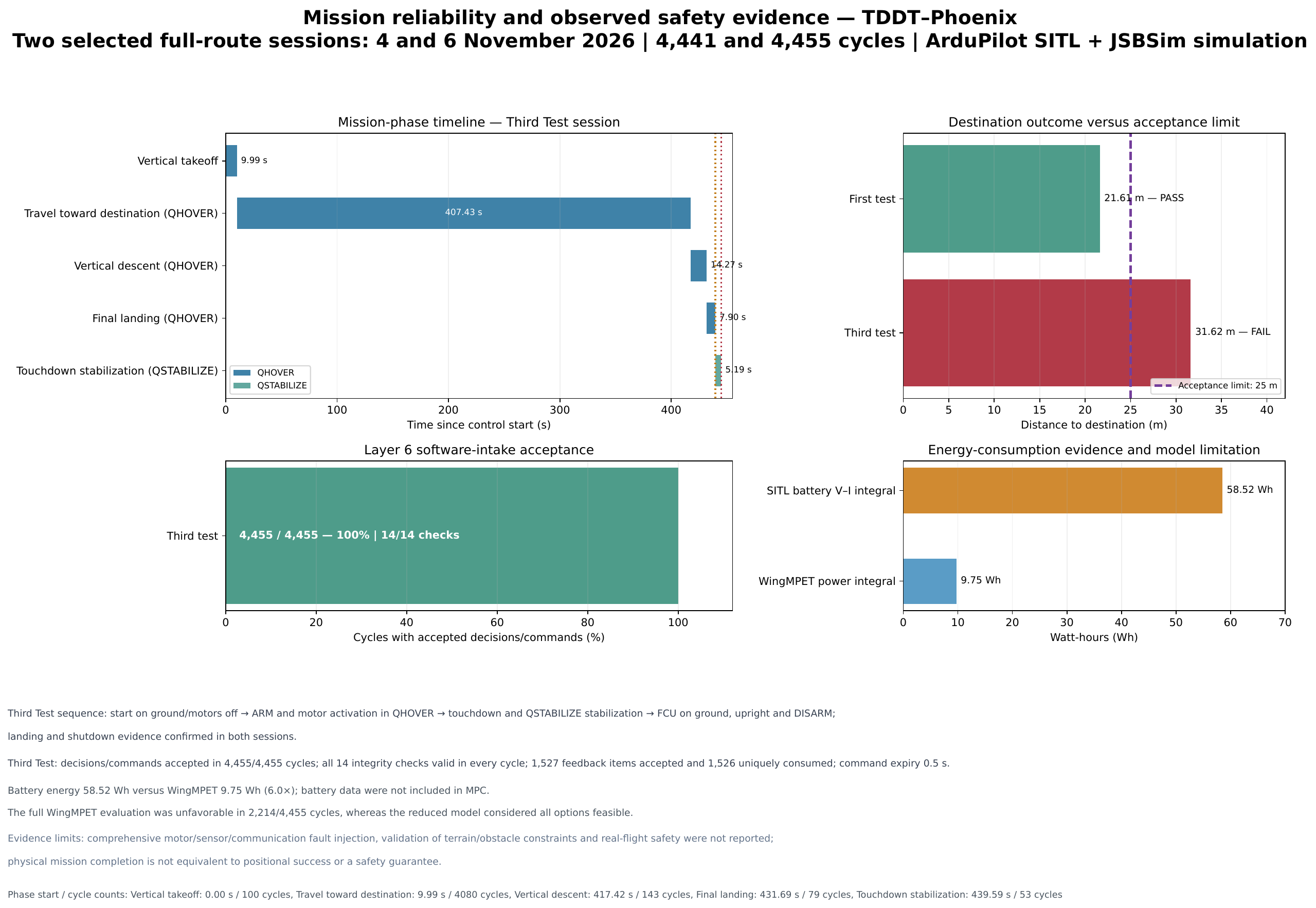}
\caption{Mission execution and safety-related evidence from two selected Native software-in-the-loop sessions, conducted on 4 and 6 November 2026. The panels summarize the five mission-phase durations in the third test, destination errors relative to the study-specific 25 m criterion, and acceptance of all 4,455 recorded decisions at the Layer 6 software input. This acceptance does not independently verify the physical response to each command. For the same 444.78 s control interval, SITL \texttt{AP\_BattMonitor} telemetry indicated 58.52 Wh, whereas the WingMPET model-power integral was 9.75 Wh; neither is a calibrated hardware-energy measurement. Confirmed touchdown, ground stabilization, and DISARM demonstrate protocol-defined termination but do not independently establish destination success or real-world flight safety.}
\label{fig:mission_execution_safety-related_evidence}
\end{figure}

Contact was detected using barometric and accelerometer evidence, such as stationary conditions or a valid impact near the ground. Subsequently, neutral command, stable ground state, and \texttt{DISARM} were confirmed. Contact stabilization in the third test lasted \(5.19\,\mathrm{s}\), although the exact activation path of the detector was not reported. The second test (5 November 2026) was accepted with a destination error of \(7.27\,\mathrm{m}\). In contrast, the third test (6 November 2026) failed the study-specific \(25\,\mathrm{m}\) destination criterion, with a final error of \(31.62\,\mathrm{m}\), despite confirmed landing and \texttt{DISARM}. In the third test, 20 of the 22 validation checks passed; the two failures concerned destination accuracy and mission completion without a guidance error. Passing input validation checks does not establish actual actuator performance.

Command and state limiters were active, but WingMPET reported unfavorable physical-feasibility indicators in 2212 cycles of the second test and 2214 cycles of the third test. In both sessions, the reduced-order control candidates satisfied their separate, more limited feasibility criteria. These results indicate a disagreement between assessment models, not an equivalent number of physical accidents or a demonstrated guarantee of global feasibility. The battery budget was not incorporated into decision-making, and full readiness for offline control was not confirmed. Sensor/actuator failures, communication interruptions, and comprehensive emergency recovery were not tested. The evidence demonstrates controlled termination of VTOL operation in SITL but does not guarantee the safety of real-world flight. Figure \ref{fig:mission_execution_safety-related_evidence} consolidates mission-phase timing, destination-acceptance outcomes, recorded command acceptance, and energy-model discrepancies observed during the simulated missions.

\subsection{Observed Errors and Weaknesses}

The median and 95th-percentile horizontal errors increased from \(2.57\) and \(3.93\,\mathrm{m}\) in the accepted second test to \(8.33\) and \(26.91\,\mathrm{m}\) in the rejected third test. Nevertheless, the 95th percentile of the prediction residual relative to the sensor-based estimate was only \(0.0134\,\mathrm{m}\); therefore, internal consistency does not imply absolute positioning accuracy. The absence of an independent horizontal measurement and an anemometer limited drift correction.

In the second and third tests, 55 and 74 cycles exceeded the nominal \(100\,\mathrm{ms}\) period, and maximum processing times reached \(171.63\) and \(181.71\,\mathrm{ms}\), respectively; no independent control consequence was established. In the third test, WingMPET reported unfavorable physical-feasibility indicators in 2,214 cycles, although the reduced-order control candidates satisfied their more limited criteria. Over the same recorded 444.78 s control interval, integration of SITL \texttt{AP\_BattMonitor} voltage–current telemetry yielded \(58.52\,\mathrm{Wh}\), compared with \(9.75\,\mathrm{Wh}\) obtained by integrating WingMPET model power. The approximately sixfold discrepancy reflects differences between two simulation-based energy models, not an error measured against physical hardware. The SITL battery monitor reported zero remaining capacity during the run, while the WingMPET state-of-charge input remained fixed at 1.0; battery-energy constraints were not incorporated into MPC decision-making.

\begin{figure}[htbp]
\centering
\includegraphics[width=0.99\linewidth,alt={Multi-panel comparison of observed limitations across the second and third simulation tests, including destination-position discrepancies, horizontal navigation errors, processing delays, feasibility-assessment disagreements, and energy-model differences. The third test records 2,214 unfavorable WingMPET feasibility indicators and simulated energy estimates of 58.52 Wh versus 9.75 Wh.}]{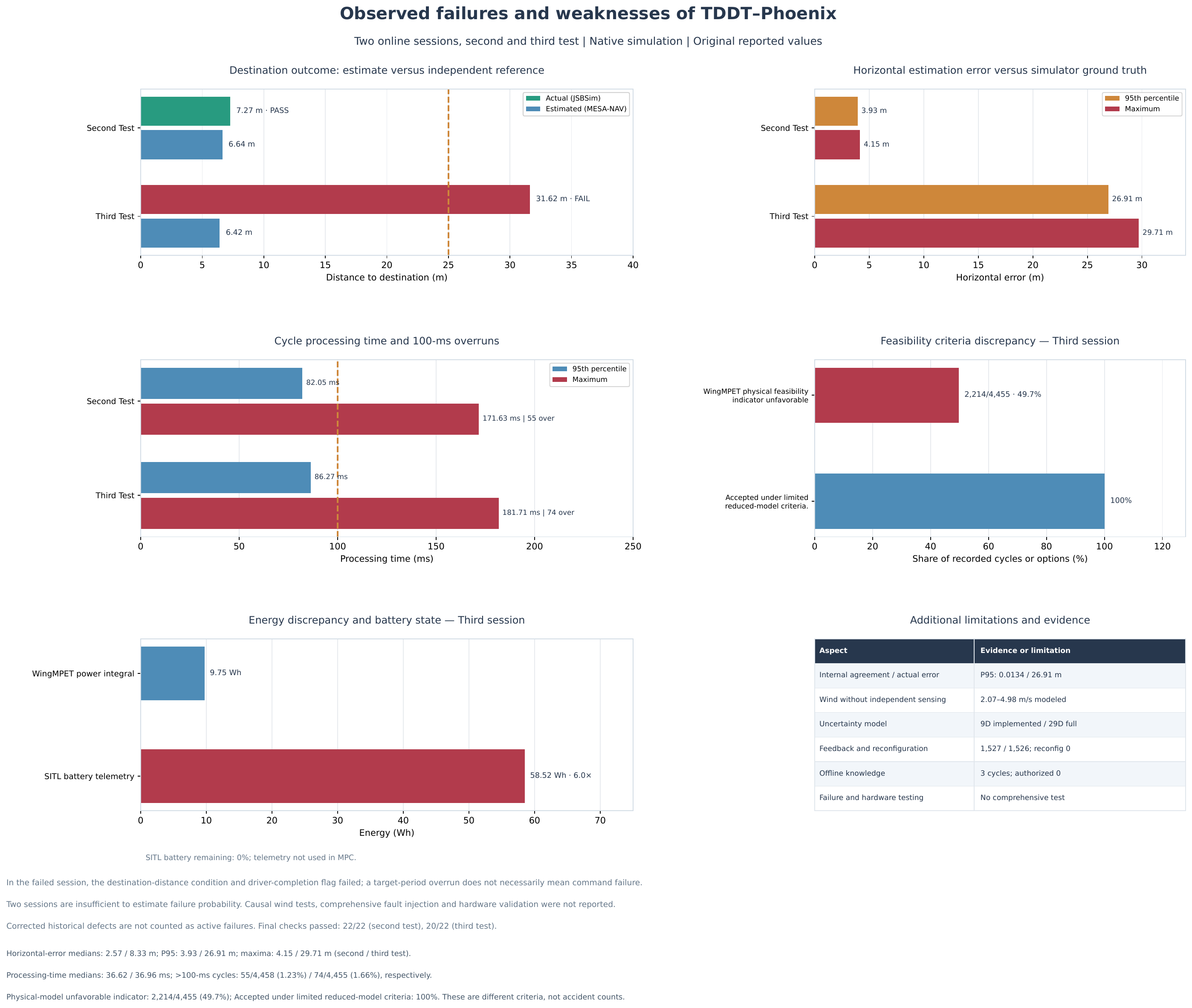}
\caption{Summary of observed limitations in the second and third Native TDDT–Phoenix sessions. The panels compare physical and estimated destination distances, horizontal position-estimation errors, computational timing and cycle overruns, and the disagreement between WingMPET physical-feasibility indicators and the reduced-order model's limited candidate-acceptance criteria. In the third test, 2,214 unfavorable WingMPET indicators were recorded; these are not counts of physical accidents. The energy panel compares the SITL battery-telemetry integral (58.52 Wh) and WingMPET model-power integral (9.75 Wh) over the same recorded control interval, without implying hardware-energy validation. Additional limitations concerning observability, uncertainty, feedback, and validation coverage are identified. The results do not establish causal effects or generalized failure probabilities.}
\label{fig:summary_observed_limitations_second_third_Native}
\end{figure}

Despite passing the 14 consistency checks for all 4455 packets in the third test, one of the 1527 accepted feedback items was not consumed. Comprehensive assessment of trust and network stability, uncertainty quantification, and executable obstacle and terrain constraints were not recorded; the causal contribution of wind was also not determined without independent measurements or tests under matched conditions. The offline report contained only three safety-hold cycles; neither an authorized final decision nor a demonstrated closed-loop effect was recorded. These limitations are discussed in Section 7. Figure \ref{fig:summary_observed_limitations_second_third_Native} consolidates the principal observed weaknesses, including positioning drift, computational delays, model-feasibility disagreements, energy-estimation discrepancies, and remaining validation limitations.

\section{Discussion and Scientific Analysis}

This section examines the operational value of TDDT, the gap between the implementation and the theoretical formalism, the role of interactions among the twins, and the requirements for practical deployment based on the experimental results.

\subsection{Architectural and Implementation Achievements}

The principal achievement is the traceable coupling of independent navigation, flight dynamics, and meteorological twins \cite{ref09_amiri2026transdomain}. TDOC connected estimation and meteorological information through the shared state to WingMPET prediction, decision-making, execution, and feedback. However, this coupling does not inherently improve navigation accuracy.

In the third test, 4,455 atomic three-twin packets passed 14 contractual checks, and predictions and decisions were recorded in every cycle. Of the 1,527 accepted feedback records, 1,526 unique records were used for limited correction of the WingMPET disturbance. The evidence confirms coordination and traceability but does not establish the physical accuracy of the models.

The second test was accepted with a destination error of \(7.27\,\mathrm{m}\). However, despite confirmed takeoff, landing, and \texttt{DISARM}, the third test exceeded the \(25\,\mathrm{m}\) threshold with an error of \(31.62\,\mathrm{m}\). The 95th-percentile horizontal error of \(26.91\,\mathrm{m}\) demonstrated that internal consistency does not guarantee absolute positioning accuracy. Furthermore, 74 cycles in the third test exceeded the nominal \(100\,\mathrm{ms}\) period; hard real-time operation was not established. The dynamic network, adaptive trust, consistency center, and offline knowledge readiness were not validated. Table~\ref{tab:tddt_independent_comparison} summarizes the integration evidence and its limits;
superiority over independent twins and the causal contributions of meteorology and feedback were not tested.

\begin{table*}[htbp]


\definecolor{TDDTNavy}{HTML}{24364D}
\definecolor{TDDTStripe}{HTML}{F0F4F9}
\definecolor{TDDTGreen}{HTML}{DDF2E9}
\definecolor{TDDTYellow}{HTML}{FFF0D8}
\definecolor{TDDTGray}{HTML}{E7ECF2}
\definecolor{TDDTRed}{HTML}{F7E1E5}
\definecolor{TDDTText}{HTML}{24364D}

\centering

\caption{Compact Comparison of the Trans-Domain Architecture
and Independent Digital Twins}

\label{tab:tddt_independent_comparison}

\vspace{2pt}

{\footnotesize
\textit{Native execution evidence and limits of inference;
Second and Third sessions.}
\par}

\vspace{6pt}

\begingroup

\footnotesize
\color{TDDTText}

\setlength{\tabcolsep}{4pt}
\renewcommand{\arraystretch}{1.3}

\begin{tabularx}{\textwidth}{
    >{\raggedright\arraybackslash}p{0.11\textwidth}
    >{\raggedright\arraybackslash}p{0.24\textwidth}
    >{\raggedright\arraybackslash}X
    >{\raggedright\arraybackslash}p{0.24\textwidth}
}

\toprule

\rowcolor{TDDTNavy}

\textcolor{white}{\textbf{Dimension}} &

\textcolor{white}{\textbf{Independent Twins;
Structural Comparison}} &

\textcolor{white}{\textbf{TDDT Integration and Key Evidence}} &

\textcolor{white}{\textbf{Defensible Conclusion}} \\

\midrule


\rowcolor{TDDTStripe}

Domain coupling &

Navigation, dynamics, and climate are computed
separately; external coupling is required. &

Three twins exchange state, wind, and density
for joint prediction.

\textbf{3 twins $\cdot$ active links} &

\cellcolor{TDDTYellow}
Coupling confirmed; the causal effect of wind
on accuracy improvement was not measured. \\

\addlinespace


Shared state &

Time, unit, and version alignment is not
inherently required. &

A time-stamped atomic packet is consumed
for prediction and decision.

\textbf{4,455 snapshots $\cdot$ 14 checks
$\cdot$ 0 mismatch} &

\cellcolor{TDDTGreen}
Contractual consistency in the active
path was confirmed. \\

\addlinespace


\rowcolor{TDDTStripe}

Coordination and execution &

Ordering models and phases, and converting
outputs into commands, requires a separate
management pathway. &

The guidance cycle, decision, command,
and simulator response are connected.

\textbf{4,455 Layer 6 intake acceptances
$\cdot$ 444.78 s}

\textbf{touchdown $\cdot$ DISARM} &

\cellcolor{TDDTGreen}
The loop for the tested VTOL phases
was executed. \\

\addlinespace


Prediction and decision &

Domain predictions are not necessarily
used in control. &

The reduced model and limited predictive-control
selection are connected to the loop.

\textbf{39,039 candidates
(4,323 $\times$ 9 + 132 $\times$ 1)}

\textbf{2,214 unfavorable WingMPET
indicators (third test)} &

\cellcolor{TDDTYellow}
Control candidates were selected and
accepted at the software input;
independent verification of every
actuator response and global physical
feasibility was not established. \\

\addlinespace


\rowcolor{TDDTStripe}

Corrective feedback &

Model error may remain separate from
commands and observations. &

Valid feedback is linked to the corresponding
prediction and command.

\textbf{1,527 accepted
$\cdot$ 1,526 uniquely used} &

\cellcolor{TDDTYellow}
Limited flight-model correction;
its quantitative benefit has not
been established. \\

\addlinespace


Independent evaluation &

Agreement among internal models is not
a sufficient criterion for destination accuracy. &

Simulator ground truth is used only for
evaluation; actual error is recorded
alongside the estimate.

\textbf{Second Test: 7.27 m PASS
$\cdot$ Third Test: 31.62 m FAIL}

\textbf{P95: 0.0134 m residual /
26.91 m true error} &

\cellcolor{TDDTGreen}
Operational-loop success did not
guarantee destination accuracy. \\

\addlinespace


\rowcolor{TDDTStripe}

Time and energy &

Execution cost and consumption of each
model require a common metric. &

Cycle-time performance and the discrepancy
between energy estimation and simulator
battery telemetry are reported.

\textbf{P95: 86.27 ms
$\cdot$ 74 cycles $>$ 100 ms}

\textbf{Same 444.78 s interval:}

\textbf{9.75 Wh WingMPET /
58.52 Wh SITL battery telemetry} &

\cellcolor{TDDTYellow}
Cycle-time performance was measured,
but a comparative timing advantage
was not established; battery information
was not used in decision-making. \\

\addlinespace


Network and offline knowledge &

Changing relationships and transferring
knowledge depend on separate management. &

Network reconfiguration and prior knowledge
are defined in architectural contracts.

\textbf{1 topology $\cdot$ 0 changes}

\textbf{0 authorized offline decisions} &

\cellcolor{TDDTGray}
Network dynamism and offline-control
readiness were not confirmed. \\

\addlinespace


\rowcolor{TDDTStripe}

Advantage over independent twins &

A fair comparison requires a baseline
under matching conditions. &

A coordinated structure and data trace
are available for assessing the contributions
of individual components.

\textbf{No controlled baseline
$\cdot$ No ablation} &

\cellcolor{TDDTRed}
Quantitative or causal superiority
has not been demonstrated. \\

\bottomrule

\end{tabularx}

\endgroup

\vspace{5pt}


\begin{minipage}{\textwidth}

\scriptsize

\textbf{Color legend:}

\colorbox{TDDTGreen}{\strut Green}:
Execution evidence within the simulation scope
\quad
\colorbox{TDDTYellow}{\strut Yellow}:
Limited execution

\colorbox{TDDTGray}{\strut Gray}:
Theoretical definition or unconfirmed readiness
\quad
\colorbox{TDDTRed}{\strut Red}:
Untested comparative claim

\vspace{4pt}

\textit{Independent twins were not executed as
a baseline under matched test conditions;
this column provides a structural comparison.
All reported physical evidence was obtained
from the simulation environment.}

\end{minipage}

\end{table*}

\subsection{Comparison of the Implementation with the Theoretical Formalism}

The Native implementation represents a limited realization of the seven-layer architecture rather than a complete implementation of Dynamic N-Twin. Table~\ref{tab:tddt_formalism_native_comparison} presents the identified gaps.

\begin{table*}[htbp]


\definecolor{TDDTNavy}{HTML}{24364D}
\definecolor{TDDTStripe}{HTML}{F0F4F9}
\definecolor{TDDTGreen}{HTML}{DDF2E9}
\definecolor{TDDTYellow}{HTML}{FFF0D8}
\definecolor{TDDTGray}{HTML}{E7ECF2}
\definecolor{TDDTRed}{HTML}{F7E1E5}
\definecolor{TDDTText}{HTML}{24364D}

\centering

\caption{Component-Level Comparison of the TDDT Formalism
and Native Implementation}

\label{tab:tddt_formalism_native_comparison}

\vspace{6pt}

\begingroup

\footnotesize
\color{TDDTText}

\setlength{\tabcolsep}{4pt}
\renewcommand{\arraystretch}{1.3}

\begin{tabularx}{\textwidth}{
    >{\raggedright\arraybackslash}p{0.18\textwidth}
    >{\raggedright\arraybackslash}X
    >{\raggedright\arraybackslash}X
}

\toprule


\rowcolor{TDDTNavy}

\textcolor{white}{\textbf{Component}} &

\textcolor{white}{\textbf{Complete Formalism}} &

\textcolor{white}{\textbf{Native Implementation}} \\

\midrule


\rowcolor{TDDTStripe}

MESA-NAV &

29-dimensional error state; multi-field anchoring &

Inertial propagation; heading and altitude correction;
9-dimensional complementary covariance \\

\addlinespace


Layer 3 network &

Dynamic topology and trust; consistency center; stability &

Atomic packets; fixed directed relationships \\

\addlinespace


\rowcolor{TDDTStripe}

L-EPAC &

Prediction of error amplification and containment
of error propagation &

Time-aligned residuals; limited model correction \\

\addlinespace


Prediction and control &

Multiobjective, robust, and uncertainty-constrained &

Limited options; scalar cost; execution-level controls \\

\addlinespace


\rowcolor{TDDTStripe}

WingMPET and MicroMet &

Comprehensive path--time--energy modeling;
multivariate meteorology &

Current dynamics and wind/density;
incomplete energy and meteorological coverage \\

\addlinespace


TDOC and offline knowledge &

Multirate network policy; learning and adaptation &

Cycle orchestration; limited offline prior knowledge \\

\bottomrule

\end{tabularx}

\endgroup

\vspace{5pt}


\begin{minipage}{\textwidth}

\scriptsize

\textit{Note: L-EPAC denotes Lightweight Error Propagation,
Amplification and Containment
\cite{ref09_amiri2026transdomain}.}

\end{minipage}

\end{table*}

In MESA-NAV, magnetic and wind--ground anchoring, revisitation, and multi-hypothesis management were not active. The magnetometer primarily corrected heading, while the barometer corrected altitude. However, the pitot sensor and prior wind information did not provide independent horizontal positioning. Instead of dynamic trust weights and a consistency center, the Native network employed fixed relationships. L-EPAC\footnote{Lightweight Error Propagation, Amplification and Containment}\cite{ref09_amiri2026transdomain} is defined for predicting and containing error propagation \cite{ref09_amiri2026transdomain}, but feedback corrected only the WingMPET disturbance using a velocity residual. Consequently, shared position errors may remain undetected.

In the control process, robust multiobjective optimization was replaced by at most nine local options with a two-second horizon and a scalar cost. Nevertheless, the execution constraints do not guarantee global stability or recursive feasibility. Across 2,214 cycles, the WingMPET feasibility assessment was inconsistent with reduced-model acceptance, without establishing that an incident occurred. Comprehensive energy optimization, obstacle and terrain constraints, fixed-wing transition, and offline control learning were also not validated. Assessing the added value of these capabilities requires controlled tests.

\subsection{Role of Interactions Among Specialized Twins}

The relationships are directed and asymmetric: WingMPET receives the estimated MESA-NAV state, MicroMet wind and density at the requested location and time, and actuator outputs. Layer 3 aligns these exchanges and makes them traceable without duplicating computations.

Wind and ground speed affect relative airspeed, forces, and predicted accelerations. The initial WingMPET response provides the basis for reduced-model prediction and option selection. Its influence on decision-making was recorded, but a complete solution of the dynamics or guaranteed satisfaction of physical constraints was not established.

The estimated position determines the meteorological query. MicroMet wind serves as a navigation prior rather than an independent wind sensor or positioning anchor. Drift may distort the selected meteorological field and subsequent predictions; the magnitude of this effect was not measured. After matching commands with sensor-based observations, Layer 7 corrects the WingMPET disturbance using a velocity residual. MESA-NAV is not reset, MicroMet is not corrected, and the source of the discrepancy remains unknown. Therefore, agreement among models sharing common errors does not establish positioning accuracy.

With a fixed topology and without dynamic trust weights, a consistency center, or paired ablation tests of inter-twin coupling, wind, or feedback, no quantitative advantage in accuracy, energy, or mission success was demonstrated.

\subsection{Scalability and Requirements for Practical Deployment}

The independence of the twins and their exchange contracts facilitate hardware deployment of TDDT, but SITL does not establish deployment readiness. The flight controller remains responsible for stability and safety-critical protection. Edge execution requires recompilation of FMUs and libraries, version compatibility, and numerical correctness.

Commands pass through the graphical interface and local communication. Therefore, the hardware interface, command identity and freshness, response acknowledgment, and communication loss must be tested. Worst-case response time, end-to-end latency, memory usage, thermal conditions, task contention, and separation of reporting operations must be evaluated. SITL timing is not guaranteed on an ARM processor.

The IMU, magnetometer, barometer, and pitot sensor require testing for synchronization, calibration, vibration, propeller airflow, magnetic interference, temperature, and bias. An independent wind sensor was not active in Native. Horizontal observability and valid environmental anchors are essential for GNSS-free navigation. Equipment mass and power requirements affect the center of gravity, inertia, thrust, and battery. WingMPET, MicroMet, the motors, elevons, and command mapping require calibration against the actual system. Energy, terrain, obstacles, and fixed-wing transition must also be tested.

An independent watchdog, detection of sensor and communication failures, rejection of expired commands, and safe recovery or shutdown are necessary. Software-based shutdown alone does not guarantee safety. Validation must progress from the edge processor and HIL\footnote[8]{\textbf{HIL (Hardware-in-the-Loop):} A method for testing and validating a control system by connecting actual hardware to a simulator of physical behavior, without requiring full operation of the system in a real environment.} to ground testing, supervised flight, and fault scenarios. None of these have been validated.

\section{Research Limitations and Threats to Validity}

This section examines the limitations of the study concerning navigation, simulation, experimentation, and safety.

\subsection{Navigation and Observability Limitations}

MESA-NAV estimates relative displacement from a known origin. Without a georeferenced anchor, inertial drift and orientation errors accumulate, and a unique absolute position cannot be recovered.

The magnetometer constrains heading, the barometer constrains relative altitude, and the Pitot tube constrains airspeed; however, without an independent wind measurement, ground speed remains unknown. MicroMet wind data do not constitute an independent observation; the three-dimensional anemometer was inactive, and multi-field anchoring was not fully implemented. Bias, noise, and environmental variations introduce drift, and calibration does not guarantee its containment.

The complementary covariance does not account for all correlations, and dimensionality reduction alone does not establish statistical overconfidence. A small residual can coexist with a large actual error, while horizontal position error does not identify its causal origin. A common vertical datum was not verified; vertical and three-dimensional errors were not evaluated, and absolute error bounds, statistical integrity, and the reliability of reaching the destination were not established.

\subsection{Gap Between Simulation and Reality}

The results are limited to SITL. The WingMPET, reduced-order control, and JSBSim models differ, and the informational independence of simulator ground truth does not imply aerodynamic validity. The Phoenix coefficients are primarily based on hover conditions, while the fixed-wing coefficients have low confidence; transitions, stall behavior, and actual forces were not tested.

Virtual sensors do not fully reproduce the effects of temperature, vibration, installation, and electromagnetic interference. Baseline IMU noise was enabled, whereas artificial gyroscope drift was disabled. Servo output does not provide an independent measurement of Elevon or motor behavior; ESC\footnote{Electronic Speed Controller} delays, saturation, voltage drops, and mechanical responses were not evaluated.

MicroMet does not include a complete gust and turbulence model. In JSBSim, only the application of horizontal wind was verified; equivalence regarding vertical wind, temperature, pressure, and density was not established. No matched-condition, wind-free test was conducted. The energy discrepancy between WingMPET and the SITL battery was model-dependent, and sensor--actuator delays and actual system responses were not tested.

\subsection{Limitations of Experimental Coverage}

The protocols for diagnostic testing, hovering, and full-mission execution were not identical. Continuous telemetry does not constitute independent replication, and a unit test cannot substitute for a mission-level test. GNSS-denied operation was evaluated by disabling real and simulated GNSS positioning inputs and absolute horizontal position aiding in SITL; physical radio-frequency jamming or spoofing was not experimentally tested. Experimental coverage was limited to a single Phoenix UAV configuration, geographic region, reference route, and VTOL flight regime; fixed-wing transition, active obstacles, and variations in altitude and terrain morphology were not tested.

The meteorological component relied on wind forecasts and a limited, version-controlled multiyear regional history. However, paired online tests under wind-free, headwind, tailwind, crosswind, gust, and turbulence conditions were not conducted. The three-sample offline test demonstrated only WingMPET sensitivity; closed-loop robustness and the contribution of MicroMet were not established.

Without matched-condition comparisons and controlled ablation of the twins, coupling mechanisms, and feedback, the causal contributions and superiority of the architecture remain unknown. Autocorrelation, the limited number of independent missions, and developmental changes constrain the estimation of between-test variance, confidence intervals, and success probabilities. Offline evidence does not establish generalizability across multiple scenarios.

\subsection{Safety and Generalizability Limitations}

Verification of command provenance and freshness, ARM status, contact, and DISARM status demonstrated only protocol-defined termination; physical safety was not established. Reduced-order control with MPC/PSF\footnote{Predictive Safety Filter}/RTA labels and motion constraints does not guarantee global stability, recursive feasibility, or collision avoidance. Terrain, obstacle, and energy constraints were not certified. Position drift reduces the validity of Geofence boundaries and landing-site estimates, while ground contact does not establish landing safety.

Rejecting an invalid or expired command does not constitute safe recovery. DEGRADED, FAILSAFE, and EMERGENCY remain theoretical; sensor, motor, communication, and processor failures, as well as independent Watchdog operation and emergency recovery, were not tested. Packet consistency does not guarantee network stability or error-propagation containment.

Validity is limited to the Phoenix tail-sitter, the software/FMU versions, sensors, origin, route, meteorological conditions, and VTOL operation. Long-duration, multi-destination, and obstacle-rich missions, as well as other aircraft, require model and actuator revalidation. Failure testing, independent replication, HIL testing, and real-world flight tests are necessary; hazard rates and generalizable reliability remain unknown.

\section{Conclusion and Future Research Directions}

This section summarizes the validated contributions and future research directions.

\subsection{Main Findings and Contributions}

The main contribution is the limited implementation of the seven-layer TDDT architecture for guiding the Phoenix VTOL without online GNSS. TDOC linked the outputs of MESA-NAV and MicroMet to WingMPET predictions and control decisions; however, a quantitative advantage over independent digital twins was not demonstrated.

In ArduPilot/JSBSim, the cycle of estimation, shared state, prediction, decision-making, execution, and feedback was completed. Takeoff, VTOL movement, landing, touchdown, and DISARM were verified. The third test comprised 4,455 cycles. Of the 1,527 accepted feedback items, 1,526 unique items were used for limited disturbance correction in WingMPET; position was not corrected, and comprehensive learning was not performed.

All 4,455 atomic packets passed 14 contractual checks covering time, identity, and consistency. However, the topology remained fixed, and dynamic trust and the consistency center were not implemented. Contractual consistency does not establish physical correctness.

The second test was accepted with a destination error of \(7.27\,\mathrm{m}\). The third test, despite a valid landing, failed with a destination error of \(31.62\,\mathrm{m}\) and a horizontal RMSE of \(13.66\,\mathrm{m}\). Therefore, the closed-loop system did not guarantee drift containment or absolute positioning.

At a target rate of \(10\,\mathrm{Hz}\), the 95th percentile processing time in the third test was \(86.27\,\mathrm{ms}\); however, 74 cycles exceeded \(100\,\mathrm{ms}\). The results support lightweight, traceable predictive guidance in SITL but do not establish hard real-time performance, guaranteed safety, or a complete Dynamic N-Twin.

\subsection{Future Research Directions}

The priority is to improve the horizontal observability of MESA-NAV through independent GeoMagTwin and TerrainTwin anchors and validated maps. PressureTwin and SolarTwin provide complementary altitude and heading references, respectively, while RadioTwin depends on ambient signals. The evaluation criteria are independently assessed reductions in position and destination errors and the statistical validity of uncertainty estimates \cite{ref09_amiri2026transdomain}.

Expanding the set of digital twins is justified only when an independent model, valid data, and measurable benefits are available \cite{ref09_amiri2026transdomain}.

The proposed options are listed in Table~\ref{tab:proposed_specialized_twins}; none have been implemented or validated.

\begin{table*}[!tp]


\definecolor{TDDTNavy}{HTML}{24364D}
\definecolor{TDDTStripe}{HTML}{F0F4F9}
\definecolor{TDDTGreen}{HTML}{DDF2E9}
\definecolor{TDDTYellow}{HTML}{FFF0D8}
\definecolor{TDDTGray}{HTML}{E7ECF2}
\definecolor{TDDTRed}{HTML}{F7E1E5}
\definecolor{TDDTText}{HTML}{24364D}

\centering

\caption{Proposed Specialized Digital Twins
and Their Conditions for Effective Use}

\label{tab:proposed_specialized_twins}

\vspace{6pt}

\begingroup

\footnotesize
\color{TDDTText}

\setlength{\tabcolsep}{4pt}
\renewcommand{\arraystretch}{1.3}

\begin{tabularx}{\textwidth}{
    >{\raggedright\arraybackslash}p{0.18\textwidth}
    >{\raggedright\arraybackslash}X
    >{\raggedright\arraybackslash}X
}

\toprule


\rowcolor{TDDTNavy}

\textcolor{white}{\textbf{Digital Twin}} &

\textcolor{white}{\textbf{Research Function}} &

\textcolor{white}{\textbf{Primary Condition for Effectiveness}} \\

\midrule


\rowcolor{TDDTStripe}

GeoMagTwin &

Using spatial magnetic anomalies for position anchoring &

A validated map and a distinctive magnetic field \\

\addlinespace


TerrainTwin &

Matching elevation and terrain patterns for position
correction and obstacle assessment &

Independent observation of height above ground \\

\addlinespace


\rowcolor{TDDTStripe}

PressureTwin &

Correcting altitude and conditionally using spatial
pressure patterns &

Distinguishing pressure variations caused by
position from those caused by time \\

\addlinespace


SolarTwin &

Providing an independent solar heading reference &

Valid observation of light or polarization and
accurate timing \\

\addlinespace


\rowcolor{TDDTStripe}

RadioTwin &

Auxiliary positioning using ambient signals &

Existing transmitters and a validated map \\

\addlinespace


AirDataTwin &

Improving airspeed estimation and estimation of
aerodynamic conditions &

Reliable airflow data \\

\addlinespace


\rowcolor{TDDTStripe}

PropulsionTwin &

Modeling thrust, actuators, and execution delays &

Measurements of actual motor and control-surface
responses \\

\addlinespace


EnergyTwin &

Estimating remaining energy and mission
completion feasibility &

A validated battery model and reliable telemetry \\

\addlinespace


\rowcolor{TDDTStripe}

SensorHealthTwin &

Detecting sensor quality degradation, bias,
and faults &

Independent sensor health testing \\

\addlinespace


LandingTwin &

Predicting descent, touchdown, and landing
conditions &

Reliable near-ground data \\

\addlinespace


\rowcolor{TDDTStripe}

FFD-WindTwin &

Predicting local variations in the wind field &

Flow-model validity and computation time \\

\addlinespace


FFD-AeroTwin &

Examining flow around the fuselage, wing,
and propeller &

Agreement with validated aerodynamic data \\

\addlinespace


\rowcolor{TDDTStripe}

MissionTwin &

Modeling mission constraints, phases,
and success criteria &

Separating the mission model from TDOC's
orchestration ownership \\

\bottomrule

\end{tabularx}

\endgroup

\vspace{5pt}


\begin{minipage}{\textwidth}

\scriptsize

\textit{Note: FFD denotes Fast Fluid Dynamics
\cite{ref31_zuo2012reduction}.}

\end{minipage}

\end{table*}

Completing the Dynamic N-Twin requires dynamic topology and trust, a consistency center, containment of error propagation through L-EPAC, and multi-rate coordination. Furthermore, robust predictive control, physical and energy constraints, uncertainty propagation, and offline knowledge must be independently validated \cite{ref09_amiri2026transdomain}.

The observable evidence standard proposed by Xu et al. assesses the value of a capability through its actual presence, opportunity to influence outcomes, changes in executable decisions, and operational consequences. Therefore, comparisons under equivalent conditions and component ablation tests using an independent reference should clarify benefits, failures, uncertainty, and overhead. A change in command alone does not indicate improvement \cite{ref27_xu2026observable}.

Determining the valid operating envelope requires validation on an edge processor, in HIL, through ground testing, and in actual flight, with controlled faults and diverse missions \cite{ref09_amiri2026transdomain}\cite{ref27_xu2026observable}.

\FloatBarrier

\bibliographystyle{plainnat}
\bibliography{references}

@article{ref01_jarraya2025gnss,
  title   = {{GNSS}-Denied Unmanned Aerial Vehicle Navigation: Analyzing Computational Complexity, Sensor Fusion, and Localization Methodologies},
  author  = {Jarraya, Imen and Al-Batati, Abdulrahman and Kadri, Muhammad Bilal and Abdelkader, Mohamed and Ammar, Adel and Boulila, Wadii and Koubaa, Anis},
  journal = {Satellite Navigation},
  volume  = {6},
  pages   = {9},
  year    = {2025},
  note    = {Article 9},
  doi     = {10.1186/s43020-025-00162-z},
  url     = {https://doi.org/10.1186/s43020-025-00162-z}
}

@article{ref02_alqudsi2027state,
  title   = {State-of-the-Art and Future Directions in Autonomous Navigation for {UAVs} in {GNSS}-Denied Environments: A Comprehensive Review of Techniques, Architectures, and Applications},
  author  = {Alqudsi, Yunes and Alameri, Abdullah and Abdallah, Ayman M. and Ahmed, Gamil},
  journal = {Information Fusion},
  volume  = {138},
  pages   = {104714},
  year    = {2027},
  doi     = {10.1016/j.inffus.2026.104714},
  url     = {https://doi.org/10.1016/j.inffus.2026.104714}
}

@article{ref03_ye2023review,
  title   = {A Review of Small {UAV} Navigation System Based on Multisource Sensor Fusion},
  author  = {Ye, Xiaoyu and Song, Fujun and Zhang, Zongyu and Zeng, Qinghua},
  journal = {IEEE Sensors Journal},
  volume  = {23},
  number  = {17},
  pages   = {18926--18948},
  year    = {2023},
  doi     = {10.1109/JSEN.2023.3292427},
  url     = {https://doi.org/10.1109/JSEN.2023.3292427}
}

@article{ref04_lee2025mins,
  title   = {{MINS}: Efficient and Robust Multisensor-Aided Inertial Navigation System},
  author  = {Lee, Woosik and Geneva, Patrick and Chen, Chuchu and Huang, Guoquan},
  journal = {Journal of Field Robotics},
  volume  = {42},
  number  = {7},
  pages   = {3252--3284},
  year    = {2025},
  doi     = {10.1002/rob.22546},
  url     = {https://doi.org/10.1002/rob.22546}
}

@article{ref05_maharmeh2025survey,
  title   = {A Comprehensive Survey on the Integrity of Localization Systems},
  author  = {Maharmeh, Elias and Alsayed, Zayed and Nashashibi, Fawzi},
  journal = {Sensors},
  volume  = {25},
  number  = {2},
  pages   = {358},
  year    = {2025},
  doi     = {10.3390/s25020358},
  url     = {https://doi.org/10.3390/s25020358}
}

@article{ref06_aljaafreh2026integrity,
  title   = {Integrity-Aware Hierarchical Navigation for {UAVs} Under {GNSS} Spoofing and Multi-Sensor Degradation},
  author  = {Alja'afreh, Mohammad and Karime, Ali and Al Mallah, Ranwa},
  journal = {Drones},
  volume  = {10},
  number  = {10},
  pages   = {738},
  year    = {2026},
  doi     = {10.3390/drones10100738},
  url     = {https://doi.org/10.3390/drones10100738}
}

@article{ref07_zhang2026detection,
  title   = {Detection-Aware Delayed Multi-Sensor Fusion for {UAV} {GNSS} Spoofing Mitigation: A Factor-Graph-Based Framework},
  author  = {Zhang, Yupeng and Xu, Yankun and Wen, Chengxin and Liu, Chenchen and Liu, Shaoxun and Lu, Qingyang and Fu, Peng},
  journal = {Sensors},
  volume  = {26},
  number  = {21},
  pages   = {6716},
  year    = {2026},
  doi     = {10.3390/s26216716},
  url     = {https://doi.org/10.3390/s26216716}
}

@article{ref08_gil2026model,
  title   = {A Model-Based Approach for Co-Simulation-Driven Digital Twins in Robotics},
  author  = {Gil, Santiago and Badyal, Arjun and Miyazawa, Alvaro and Larsen, Peter Gorm and Cavalcanti, Ana},
  journal = {Robotics and Autonomous Systems},
  volume  = {196},
  pages   = {105240},
  year    = {2026},
  doi     = {10.1016/j.robot.2025.105240},
  url     = {https://doi.org/10.1016/j.robot.2025.105240}
}

@article{ref09_amiri2026transdomain,
  title         = {Trans-Domain Digital Twin: Conceptual Foundations, Architecture, and Research Outlook},
  author        = {Amiri, Mansoorali},
  journal       = {arXiv preprint arXiv:2607.15908},
  year          = {2026},
  eprint        = {2607.15908},
  archivePrefix = {arXiv},
  primaryClass  = {cs.SE},
  doi           = {10.48550/arXiv.2607.15908},
  url           = {https://doi.org/10.48550/arXiv.2607.15908}
}

@inproceedings{ref10_wu2019phoenix,
  title        = {The {Phoenix} Drone: An Open-Source Dual-Rotor Tail-Sitter Platform for Research and Education},
  author       = {Wu, Yilun and Du, Xintong and Duivenvoorden, Rikky and Kelly, Jonathan},
  booktitle    = {Proceedings of the IEEE International Conference on Robotics and Automation (ICRA)},
  pages        = {5330--5336},
  address      = {Montreal, Quebec, Canada},
  year         = {2019},
  organization = {IEEE},
  doi          = {10.1109/ICRA.2019.8794433},
  url          = {https://doi.org/10.1109/ICRA.2019.8794433}
}

@article{ref11_lisus2023know,
  title   = {Know What You Don't Know: Consistency in Sliding Window Filtering With Unobservable States Applied to Visual-Inertial {SLAM}},
  author  = {Lisus, Daniil and Cohen, Mitchell and Forbes, James Richard},
  journal = {IEEE Robotics and Automation Letters},
  volume  = {8},
  number  = {6},
  pages   = {3382--3389},
  year    = {2023},
  doi     = {10.1109/LRA.2023.3268043},
  url     = {https://doi.org/10.1109/LRA.2023.3268043}
}

@article{ref13_forsling2024quarter,
  title   = {A Quarter Century of Covariance Intersection: Correlations Still Unknown?},
  author  = {Forsling, Robin and Noack, Benjamin and Hendeby, Gustaf},
  journal = {IEEE Control Systems Magazine},
  volume  = {44},
  number  = {2},
  pages   = {81--105},
  year    = {2024},
  note    = {Lecture Notes},
  doi     = {10.1109/MCS.2024.3358658},
  url     = {https://doi.org/10.1109/MCS.2024.3358658}
}

@article{ref15_bajwa2024dive,
  title   = {{DIVE}: Deep Inertial-Only Velocity Aided Estimation for Quadrotors},
  author  = {Bajwa, Angad and Champagne Cossette, Charles and Shalaby, Mohammed Ayman and Forbes, James Richard},
  journal = {IEEE Robotics and Automation Letters},
  volume  = {9},
  number  = {4},
  pages   = {3728--3734},
  year    = {2024},
  doi     = {10.1109/LRA.2024.3370006},
  url     = {https://doi.org/10.1109/LRA.2024.3370006}
}

@article{ref16_yao2024gnss,
  title   = {{GNSS}-Denied Geolocalization of {UAVs} Using Terrain-Weighted Constraint Optimization},
  author  = {Yao, Fushan and Lan, Chaozhen and Wang, Longhao and Wan, Hongfa and Gao, Tian and Wei, Zijun},
  journal = {International Journal of Applied Earth Observation and Geoinformation},
  volume  = {135},
  pages   = {104277},
  year    = {2024},
  doi     = {10.1016/j.jag.2024.104277},
  url     = {https://doi.org/10.1016/j.jag.2024.104277}
}

@article{ref17_zhu2026geomagnetic,
  title   = {Development and Prospects of Multi-Domain Geomagnetic Navigation and Positioning Technologies},
  author  = {Zhu, Libo and Li, Houpu and Zhu, Bo and Liu, Bing and Ouyang, Jineng and Xu, Lei and Bian, Shaofeng and Liu, Fujiang},
  journal = {Sensors},
  volume  = {26},
  number  = {18},
  pages   = {5970},
  year    = {2026},
  doi     = {10.3390/s26185970},
  url     = {https://doi.org/10.3390/s26185970}
}

@article{ref18_deoliveira2024pitot,
  title   = {Pitot Tube Measure-Aided Air Velocity and Attitude Estimation in {GNSS} Denied Environment},
  author  = {de Oliveira, Tomas L. and van Goor, Pieter and Hamel, Tarek and Mahony, Robert and Samson, Claude},
  journal = {European Journal of Control},
  volume  = {80},
  pages   = {101070},
  year    = {2024},
  doi     = {10.1016/j.ejcon.2024.101070},
  url     = {https://doi.org/10.1016/j.ejcon.2024.101070}
}

@article{ref19_soltaninezhad2025wind,
  title   = {A Review of Methods and Challenges for Wind Measurement by Small Unmanned Aerial Vehicles},
  author  = {Soltaninezhad, Mohammadamin and Monsorno, Roberto and Tondini, Stefano},
  journal = {Meteorological Applications},
  volume  = {32},
  number  = {3},
  pages   = {e70065},
  year    = {2025},
  doi     = {10.1002/met.70065},
  url     = {https://doi.org/10.1002/met.70065}
}

@article{ref20_chen2026wind,
  title   = {Wind Estimation Based on Flight Dynamics of Unmanned Aerial Vehicle: Influencing Variables and Its Environmental Application},
  author  = {Chen, Dukun and Su, Weifeng and Jiang, Shaojie and Yang, Honglong and Zhang, Chunsheng and Jiang, Shutong and Chang, Dongyang and Liang, Yuxin and Wang, Hao and Yang, Xin and Fu, Tzung-May and Zeng, Zhenzhong and Zhu, Lei and Shen, Huizhong and Wang, Chen and Ye, Jianhuai},
  journal = {Atmospheric Chemistry and Physics},
  volume  = {26},
  pages   = {3607--3620},
  year    = {2026},
  doi     = {10.5194/acp-26-3607-2026},
  url     = {https://doi.org/10.5194/acp-26-3607-2026}
}

@article{ref21_lu2026graph,
  title   = {A Graph-Optimization-Based Tightly Coupled Multi-Source Positioning Method for {UAVs} in {GNSS}-Denied Environments},
  author  = {Lu, Jia and Zhang, Zuyin and Qi, Yishen and Song, Ping},
  journal = {Measurement},
  volume  = {280},
  pages   = {121787},
  year    = {2026},
  doi     = {10.1016/j.measurement.2026.121787},
  url     = {https://doi.org/10.1016/j.measurement.2026.121787}
}

@article{ref22_amiri2026greenhouse,
  title   = {Trans-Domain Digital Twins with Adaptive Intelligence for Optimizing Greenhouse Crop Production},
  author  = {Amiri, Mansoorali and Delcourt, Kevin and Sahraoui, Houari},
  journal = {Software and Systems Modeling},
  year    = {2026},
  note    = {Published online 21 September 2026; volume and page numbers not yet assigned in the verified publisher record},
  doi     = {10.1007/s10270-026-01434-1},
  url     = {https://doi.org/10.1007/s10270-026-01434-1}
}

@article{ref23_yang2025digital,
  title   = {Digital Twin-Based Obstacle Avoidance for Unmanned Aerial Vehicles Using Feedforward-Feedback Control},
  author  = {Yang, Hongjiu and Sun, Shaopeng and Xia, Yuanqing and Li, Peng},
  journal = {IEEE Transactions on Vehicular Technology},
  volume  = {74},
  number  = {6},
  pages   = {8721--8733},
  year    = {2025},
  doi     = {10.1109/TVT.2025.3536776},
  url     = {https://doi.org/10.1109/TVT.2025.3536776}
}

@article{ref24_kumar2026indoor,
  title   = {Digital Twin for Drone Indoor Autonomous Navigation},
  author  = {Kumar, M. Jaswanth and Singh, Satyam and Saneen, Ahnaf and Thomas, Della},
  journal = {IEEE Sensors Letters},
  volume  = {10},
  number  = {2},
  pages   = {1--4},
  eid     = {6001104},
  year    = {2026},
  doi     = {10.1109/LSENS.2025.3642424},
  url     = {https://doi.org/10.1109/LSENS.2025.3642424}
}

@article{ref25_robles2026composite,
  title   = {Composite Digital Twins: From Formalization to a Scalable, Semantic Graph-Based Architecture with {OpenTwins}},
  author  = {Robles, Julia and Infante, Sergio and Martin, Cristian and Diaz, Manuel},
  journal = {IEEE Transactions on Software Engineering},
  pages   = {1--26},
  year    = {2026},
  note    = {Early Access},
  doi     = {10.1109/TSE.2026.3727063},
  url     = {https://doi.org/10.1109/TSE.2026.3727063}
}

@misc{ref26_nn44nniri_tddt_phoenix,
  author       = {{Mansoorali Amiri}},
  title        = {{nn44nniri/TDDT\_phoenixdrone\_navigation\_GPSdenied:
                   TDDT phoenix drone navigation GPSdenied 1.0.0}},
  year         = {2026},
  publisher    = {Zenodo},
  howpublished = {Computer software},
  doi          = {10.5281/zenodo.24419538},
  url          = {https://doi.org/10.5281/zenodo.24419538}
}

@article{ref27_xu2026observable,
  title         = {When Does an Added Planning Capability Create Operational Value? An Observable Evidence Standard},
  author        = {Xu, Yihua and Lyu, Yan and Li, Jize and He, Jiani and Lin, You and Hu, Yidan and Qiang, Rubing},
  journal       = {arXiv preprint arXiv:2610.27305},
  year          = {2026},
  eprint        = {2610.27305},
  archivePrefix = {arXiv},
  primaryClass  = {cs.AI},
  doi           = {10.48550/arXiv.2610.27305},
  url           = {https://doi.org/10.48550/arXiv.2610.27305}
}

@article{ref30_sun2026eadr,
  title   = {{EADR}: Efficient Runtime Detection and Recovery of Actuator Attacks on {UAVs}},
  author  = {Sun, Cong and He, Penghao and Wang, Yunbo and Zhang, Zongxu and Wei, Xiaomin},
  journal = {Cybersecurity},
  volume  = {9},
  pages   = {63},
  year    = {2026},
  note    = {Article 63},
  doi     = {10.1186/s42400-025-00483-1},
  url     = {https://doi.org/10.1186/s42400-025-00483-1}
}

@article{ref31_zuo2012reduction,
  title={Reduction of Numerical Diffusion in FFD Model},
  author={Zuo, Wangda and Jin, Mingang and Chen, Qingyan},
  journal={Engineering Applications of Computational Fluid Mechanics},
  volume={6},
  number={2},
  pages={234--247},
  year={2012},
  publisher={Taylor \& Francis},
  doi={10.1080/19942060.2012.11015418},
  url={https://doi.org/10.1080/19942060.2012.11015418}
}

\end{document}